\documentclass[letterpaper]{article}
\usepackage[preprint]{aaai2027}
\usepackage[hyphens]{url}
\usepackage{graphicx}
\usepackage{natbib}
\usepackage{caption}
\usepackage{booktabs}
\usepackage{amsmath}
\usepackage{amssymb}
\usepackage{float}

\graphicspath{{assets/}}

\newcommand{\indicator}[1]{\mathbf{1}\!\left[#1\right]}

\title{Witness Evidence Portfolios: Single-Prefill Risk Detection for Closed Multimodal Answers}
\author{
Feixiang Liu\textsuperscript{\rm 1,2},
Shiye Wang\textsuperscript{\rm 1},
Qiang Qiu\textsuperscript{\rm 1},
Zheng Wang\textsuperscript{\rm 1}
}
\affiliations{
\textsuperscript{\rm 1}State Key Laboratory of AI Safety, Institute of Computing Technology, CAS\\
\textsuperscript{\rm 2}University of Chinese Academy of Sciences\\
liufeixiang23s@mails.ucas.ac.cn, wangshiye@ict.ac.cn, qiuqiang@ict.ac.cn,
wangzheng@ict.ac.cn
}

\begin{document}

\maketitle

\begin{abstract}
Reliable deployment of multimodal large language models (MLLMs) requires deciding whether a confident visual answer should be trusted, reviewed, or routed to a stronger system. Confidence scores capture candidate margins, but not where the estimated signed visual readouts associated with those margins come from or how they are distributed. We study inference-time risk detection for closed visual answers using the same white-box prefill path that produces the answer. Witness Evidence Portfolios (WEP) first estimates, layer by layer, which visual contributions support or contradict the predicted candidate. It summarizes these contributions through two interpretable route families: question-related evidence provenance and signed evidence concentration. Nested grouped validation chooses the more reliable family and a sparse top-$k$ route portfolio, which is fused with candidate confidence. WEP needs no image perturbation, decoding change, backward pass, or external verifier. Across three MLLMs and four binary-answer benchmarks, WEP improves mean error AP by $0.134$. All 12 model--dataset gains are positive, and image-cluster bootstrap intervals are strictly positive on 10 pairs. WEP targets white-box closed-answer systems and uses a labeled calibration slice.
\end{abstract}

\begin{center}
\textbf{Code:} \url{https://github.com/SouthWinter/WEP}
\end{center}

\section{Introduction}

Reliable deployment of multimodal large language models (MLLMs) requires more than producing a plausible visual answer \citep{rohrbach2018object,li2023pope,wang2023amber,guan2024hallusionbench,wang2025mllmsurvey}. A system also needs to decide whether the current answer should be trusted, reviewed, or routed to a stronger model. Consider a model that confidently answers ``no'' to the question ``Are there any carrots in this figure?'' even when the image contains carrots. The answer margin can be large even when its associated internal visual readouts concentrate on a chart title, background context, or another salient region rather than the visual evidence corresponding to carrots. Confidence alone cannot distinguish this unsupported answer from a genuinely grounded one. We use this carrots question as a running example below.

We study this need within reliable and selective visual answering \citep{whitehead2022reliablevqa,dancette2023selectivevqa}. Our setting is \emph{closed-answer MLLM risk detection}, and the main evaluation focuses on binary visual questions \citep{li2023pope,liu2024mmbench}. The candidate set is fixed, and a single prefill pass already provides the candidate logits. Given an image, a question, and the model's current prediction, the goal is to rank predictions by their likelihood of being incorrect so that outputs can be reviewed, abstained, routed, or audited. Candidate confidence is the natural baseline \citep{hendrycks2017baseline,guo2017calibration,geifman2017selective}, but it only measures how far the predicted candidate is from its competitors; it does not test whether signed visual readouts associated with that margin align with the question.

Existing risk and hallucination methods leave this gap open. Learned uncertainty scores and generic internal-state detectors predict correctness from probabilities, hidden states, or activations \citep{lakshminarayanan2017simple,gal2016dropout,chen2024inside,su2024mind,kogilathota2026halp,zhang2026vibprobe}, but do not constrain the source of decision evidence to the regions requested by the question. Perturbation, contrastive decoding, and verifier-based methods can expose unstable or unsupported answers \citep{leng2024vcd,wang2024icd,huang2024opera,yin2023woodpecker,favero2024m3id,zhang2026pti,wang2026taco}, but require additional image views, decoding changes, or verifier calls. Attention and grounding analyses move closer to visual evidence \citep{wang2026sameattention,zhao2025vga,hoang2026pas}, yet attention alone does not show whether a visual token moves the current candidate margin, nor whether that movement supports or contradicts the prediction.

We ask whether estimated signed visual readouts associated with the current candidate margin provide risk information beyond confidence, and whether their provenance and spatial concentration offer complementary views of unsupported decisions.

To answer these questions, we propose Witness Evidence Portfolios (WEP). WEP first estimates, layer by layer, which internal visual contributions support or contradict the predicted candidate margin. It then constructs two route families. Provenance routes compare evidence near question-bound witnesses with a uniform visual reference; concentration routes measure whether signed evidence is narrowly concentrated or spatially diffuse. Inner grouped validation chooses the reliable family, after which a sparse top-$k$ portfolio is fitted on the full calibration split and fused with candidate confidence. The resulting score uses the original candidate prefill, instrumented only with white-box hooks and sparse read-row recomputation; it needs no second model forward, image perturbation, decoding change, backward pass, or external verifier call.

Experiments across three MLLM backbones and four closed-answer benchmarks show that WEP improves error ranking over candidate confidence and concentrates errors earlier in the review queue. Transfer and evidence-shuffle controls on the predefined provenance family indicate that its gains depend on example-matched internal evidence rather than score-scale artifacts, while matched-budget learned selectors transfer less reliably. WEP's sparse, named routes make the resulting risk score inspectable at both the example and portfolio levels.

The contributions are:
\begin{itemize}
\item We introduce a signed decision-evidence view of closed-answer risk ranking: whether estimated internal visual contributions support or contradict the current candidate margin, beyond what confidence alone reveals.
\item We introduce complementary provenance and concentration decompositions of signed visual evidence, together with a nested, low-capacity selector that chooses one route family before sparse portfolio construction and confidence fusion.
\item We validate the detector across three backbones and four benchmarks, with grouped uncertainty, selective-prediction, matched-budget, transfer, shuffle, localization, and intervention analyses.
\end{itemize}

\section{Related Work}

\noindent\textbf{Hallucination mitigation and verification.} Hallucination mitigation changes evidence, decoding, or post-hoc verification. Visual Contrastive Decoding (VCD) and Instruction Contrastive Decoding (ICD) contrast original and altered inputs; HEDGE and OPERA score or modify decoding; and Woodpecker verifies and regenerates outputs \citep{leng2024vcd,wang2024icd,gautam2025hedge,huang2024opera,yin2023woodpecker}. Other systems add perturbations, sampling, or learned checking \citep{favero2024m3id,wang2026taco}. Reallocation rescales layer-wise attention heads \citep{lu2026reallocating}, while Prefill-Time Intervention steers the prefill key--value cache \citep{zhang2026pti}. These methods improve or verify answers, often through extra paths; WEP instead scores the current answer from its existing prefill.

\noindent\textbf{Internal evidence for hallucination detection.} Internal signals provide a more direct diagnostic. Same Attention, Different Truths combines visual attention with logit-lens readouts \citep{wang2026sameattention,belrose2023tuned}; VGA and PAS use grounding or preliminary-token attention \citep{zhao2025vga,hoang2026pas}; and Overthinking tracks layer-wise hypotheses \citep{shoby2026overthinking}. HALP and VIB-Probe use query/visual states or bottlenecked attention heads \citep{kogilathota2026halp,zhang2026vibprobe}, while Attention Never Lie and VADE learn attention-based detectors \citep{zhao2026attentionnever,prabhakaran2025vade}. Generic detectors instead predict hallucination from hidden states, attention outputs, or probabilities \citep{chen2024inside,su2024mind}. WEP differs in prediction unit and evidence constraint: it ranks closed candidates using signed source-to-witness readouts associated with the current margin, rather than generic activation or attention features.

\noindent\textbf{Answer-risk detection and selective routing.} A related task ranks predictions for abstention, review, or routing. Classic selective prediction uses confidence, ensembles, Bayesian dropout, or conformal scores \citep{hendrycks2017baseline,guo2017calibration,lakshminarayanan2017simple,gal2016dropout,geifman2017selective,angelopoulos2021conformal}. Reliable VQA learns a multimodal correctness selector; Learning from Your Peers uses peer correctness under shift; and black-box selective VQA estimates neighborhood consistency with a proxy model \citep{whitehead2022reliablevqa,dancette2023selectivevqa,khan2024consistency}. Other selectors use Bayesian modeling or retrieval memory \citep{wieczorek2026variationalvqa,sarkar2026mapapsp}. WEP instead derives a complementary white-box score from the fixed backbone's original candidate path. This responds to the attribution caveat that attention and saliency may localize behavior without identifying decision support \citep{jain2019attention,wiegreffe2019attention,selvaraju2017gradcam,sundararajan2017axiomatic,abnar2020quantifying,chefer2021transformer}. WEP augments selective-routing scores with signed evidence and question alignment rather than replacing their risk-control layer.

\section{Method}

\begin{figure*}[t]
\centering
\includegraphics[width=\textwidth]{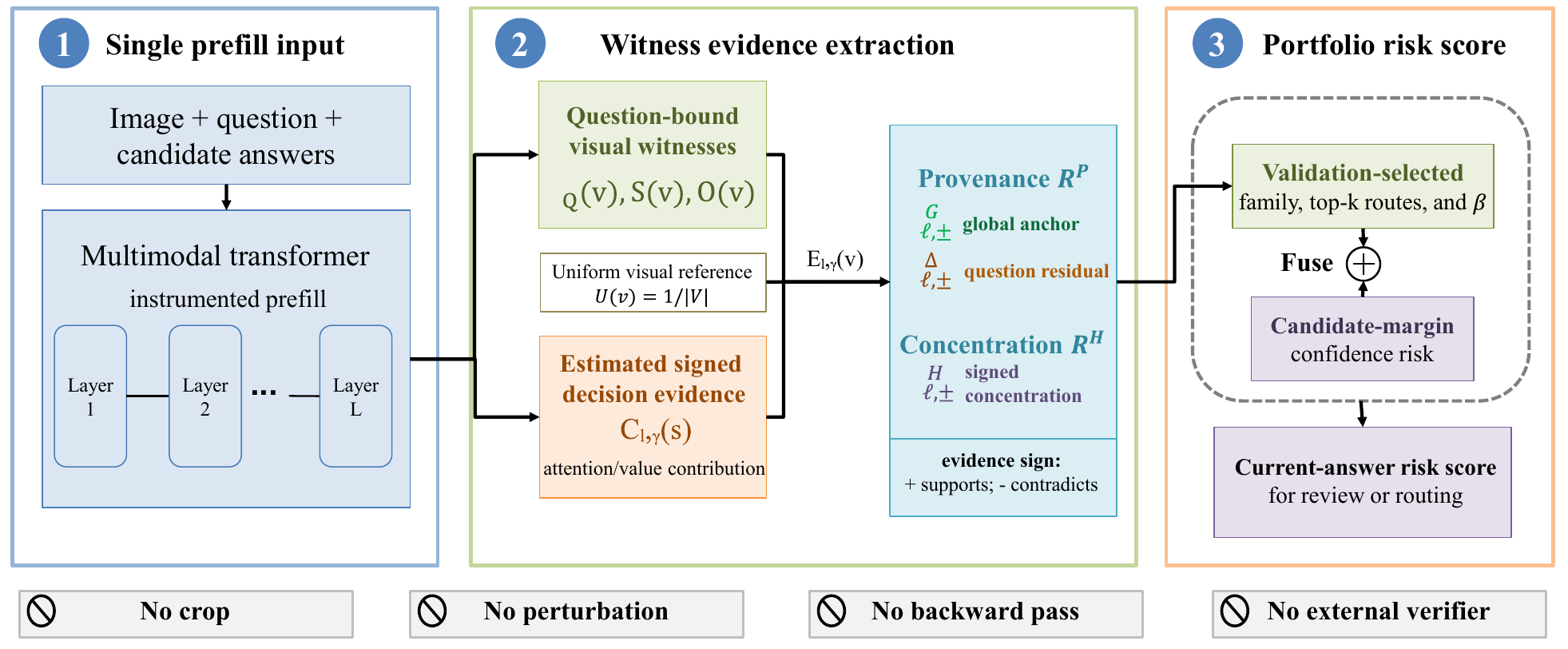}
\caption{Overview of Witness Evidence Portfolios (WEP). One instrumented prefill path supplies candidate logits and sparse internal readouts. Question-bound witnesses and estimated margin contributions form complementary provenance and signed-concentration route families. Nested grouped validation chooses one family and a sparse top-$k$ portfolio, which is fused with confidence risk to rank current predictions by error risk.}
\label{fig:method}
\end{figure*}

\subsection{Framework Overview}

WEP estimates signed internal visual evidence associated with the predicted candidate margin. Figure~\ref{fig:method} summarizes the pipeline. Decision-time attention/value readouts estimate signed source-token contributions and transport them into image space, while question tokens define visual witness maps. WEP summarizes each layer and sign in two ways. The \emph{provenance family} measures global visual overlap and question-specific departures from that reference. The \emph{concentration family} measures how narrowly the signed evidence is distributed over visual tokens. Nested validation chooses one family; a sparse portfolio of its layer/sign routes is then fused with candidate-margin risk. Conceptually, WEP proceeds from regions $(P_Q,S,O,U)$ to source contributions $C$, image-space evidence $E$, a validated route family, selected route scalars, and the final risk score.

\noindent\textbf{Running intuition.} For the carrots example, WEP estimates which tokens increase or decrease the model's predicted ``no'' margin and which visual tokens the question word ``carrots'' binds to. Provenance routes ask whether this evidence reaches the carrot witnesses rather than generic image regions. Concentration routes ask whether supportive evidence is narrowly concentrated or contradictory evidence is diffusely distributed. Validation retains the view that predicts errors reliably for the calibrated model and task.

\subsection{Problem Formulation}

An input consists of an image $I$, a question $q$, and a closed candidate set $\mathcal{Y}$. A normal prefill pass returns logits $z_y$ for $y\in\mathcal{Y}$ and prediction $\hat{y}=\arg\max_y z_y$. We assign a risk score $r(x)$ to the current prediction. The evaluation label is $e(x)=\indicator{\hat{y}\ne y^\star}$, where $y^\star$ is the benchmark label. The confidence baseline is the negative candidate margin:
\begin{equation}
r_{\mathrm{conf}}(x)=-
\left(z_{\hat{y}}-\max_{y\ne\hat{y}}z_y\right).
\end{equation}
Larger scores mean higher error risk.

\subsection{Question-Bound Witness Maps}

Let $V$ be visual token positions and $Q$ be the question token span. The witness map should emphasize regions bound to the question rather than all salient tokens. From the prefill pass, we recompute sparse attention/value read rows for question tokens and the decision position. For question token $j$, $B_j(v)$ is its normalized binding distribution over visual token $v$. Token importance $\omega_j$ is estimated by how much the decision position reads from $j$, with $\sum_{j\in Q}\omega_j=1$. The question witness map is
\begin{equation}
P_Q(v)=\sum_{j\in Q}\omega_j B_j(v).
\label{eq:witness}
\end{equation}
In the running example, $B_{\text{carrots}}(v)$ is the visual distribution attached to the token ``carrots'', and $P_Q$ aggregates it with the rest of the question according to each token's decision relevance. A rule-based parser identifies subject/object spans directly in the visible question; entity maps $S(v)$ and $O(v)$ then subtract a control-token visual sink prior. If parsing fails, both maps fall back to $P_Q(v)$. When $S=O=P_Q$, the support minimum and contradiction maximum below reduce to the same full-question overlap; the two-entity rule is active only when paired entity spans are available. We also define a question-independent visual reference $U(v)=1/|V|$. It is not a witness map: it controls for how the same signed evidence overlaps the visual field before question-specific alignment is credited.

\subsection{Estimated Decision Evidence}

The witness map identifies where the question is visually bound; it does not by itself determine what evidence moves the answer. To measure decision evidence, we define for each candidate $y$ a one-vs-competitors readout direction:
\begin{equation}
d_y=W_{\mathrm{out}}[y]-\sum_{y'\ne y}p(y'\mid\mathcal{Y}\setminus y)W_{\mathrm{out}}[y'] ,
\end{equation}
where $W_{\mathrm{out}}$ is the unembedding matrix and competitor weights are a softmax over candidate logits excluding $y$. Projecting internal values onto $d_y$ estimates whether a source token increases or decreases that margin, following logit-level readouts and circuit decompositions \citep{elhage2021mathematical,belrose2023tuned,riggs2024logitprisms}. At layer $\ell$, the contribution of source token $s$ to candidate $y$ at decision position $t$ is
\begin{equation}
C_{\ell,y}(s)=
\sum_h
\left\langle
\tilde d_y,
W_O^{\ell,h}\left(A_{\ell,h}(t,s)V_{\ell,h}(s)\right)
\right\rangle .
\label{eq:contribution}
\end{equation}
Here $A_{\ell,h}(t,s)$ is the attention weight from the decision position to source token $s$, $V_{\ell,h}(s)$ is the value vector, and $W_O^{\ell,h}$ is the head output projection. For pre-normalization decision state $x_t$, the local readout is computed analytically as $\tilde d_y=\nabla_{x_t}\langle d_y,\operatorname{RMSNorm}(x_t)\rangle$; it does not backpropagate through the network. For the predicted answer ``no'', $C_{\ell,\text{no}}(s)>0$ means that source token $s$ pushes the layer readout toward ``no'', even if $s$ is a chart title or background token rather than a carrot witness. Because the attention update is additive over source tokens, Equation~\eqref{eq:contribution} obeys the exact first-order completeness identity
\begin{equation}
\sum_s C_{\ell,y}(s)
=\langle\tilde d_y,\Delta x_{\ell,t}^{\mathrm{attn},0}\rangle ,
\label{eq:local-completeness}
\end{equation}
where $\Delta x_{\ell,t}^{\mathrm{attn},0}$ is the bias-free attention update. This identity decomposes a local directional derivative, not the complete nonlinear network. An all-layer audit verifies numerical closure and agreement with exact same-sublayer margin changes across all three backbones; separate end-to-end patching tests whether the identified tokens affect final logits.

Source-token contributions are transported to visual witness space so they can be compared with $P_Q,S,$ and $O$. A visual source token maps to itself, while a question source token maps through its own visual binding distribution:
\begin{equation}
E_{\ell,y}(v)=\bar C_{\ell,y}(v)+
\sum_{j\in Q}\rho_j\bar C_{\ell,y}(j)B_j(v).
\label{eq:transport}
\end{equation}
Here $\bar C$ is $C$ normalized by its source-token $\ell_1$ mass and $\rho_j\in[0,1]$ is the visually sourced fraction of question token $j$. Specifically, if $M_\ell(a,b)$ denotes the hook's attention/value read mass, $\rho_j=\sum_{\ell,v\in V}M_\ell(j,v)/ \max(\sum_{\ell,s}M_\ell(j,s),\epsilon)$. Other text mass maps to a null bucket and is excluded from visual evidence overlap. This produces estimated signed visual evidence $E_{\ell,y}(v)$ for each layer and candidate. Thus $E_{\ell,\text{no}}(v)$ asks where the internal evidence for ``no'' lands after text-mediated contributions are transported back to visual space.

\subsection{Signed Route Families}

For the predicted answer $\hat y$, each layer first yields question-bound support and contradiction scores. Support should cover both queried entities, since one-sided evidence can spuriously favor a relation or attribute answer. Contradiction is risky if it overlaps either entity, because evidence near a requested witness is pushing against the prediction. The question-bound scores are
\begin{equation}
\begin{aligned}
g^{Q,+}_{\ell}
&=\kappa_{\hat y}\min(a^S_{\ell,+},a^O_{\ell,+}),\\
a^S_{\ell,+}
&=\sum_v \min(E^+_{\ell,\hat y}(v),S(v)),\\
a^O_{\ell,+}
&=\sum_v \min(E^+_{\ell,\hat y}(v),O(v)).
\end{aligned}
\label{eq:support}
\end{equation}
and
\begin{equation}
\begin{aligned}
g^{Q,-}_{\ell}
&=\kappa_{\hat y}\max(a^S_{\ell,-},a^O_{\ell,-}),\\
a^S_{\ell,-}
&=\sum_v \min(E^-_{\ell,\hat y}(v),S(v)),\\
a^O_{\ell,-}
&=\sum_v \min(E^-_{\ell,\hat y}(v),O(v)).
\end{aligned}
\label{eq:contradiction}
\end{equation}
Here $E^+$ and $E^-$ are separately normalized positive and negative evidence. The candidate-level factor is $\kappa_y=\sum_\ell\gamma_{\ell,y}\sum_s|\bar C_{\ell,y}(s)| \sum_{v\in V}\Pi_s(v)$, where $\Pi_s$ is the source-to-visual transport and $\gamma_{\ell,y}$ is normalized absolute layer-margin change. It is shared across layers; validation selects informative layers instead of applying a hand-designed per-layer margin gate. The $\min$ terms are histogram-intersection overlaps.

To distinguish question-specific alignment from generic visual evidence geometry, we recompute the same two scores with $S=O=U$, obtaining $g^{G,+}_{\ell}$ and $g^{G,-}_{\ell}$. We convert both score families to risk orientation and center the question routes by the global reference:
\begin{equation}
\begin{aligned}
r^b_{\ell,+}&=-g^{b,+}_{\ell},&
r^b_{\ell,-}&= g^{b,-}_{\ell},\quad b\in\{Q,G\},\\
r^{\Delta}_{\ell,s}&=r^Q_{\ell,s}-r^G_{\ell,s},&
\mathcal R&=\{r^G_{\ell,s},r^{\Delta}_{\ell,s}\}_{\ell,s\in\{+,-\}}.
\end{aligned}
\label{eq:reference-routes}
\end{equation}
Thus $r^G$ records evidence overlap against the question-independent uniform reference; the route itself remains conditioned on the current image--question path through $E$. The residual $r^\Delta$ records excess witness alignment relative to that reference. In the carrots case, the residual exposes whether evidence aligns with carrot witnesses more than with the visual field in general. These routes form the provenance family $\mathcal R^{P}=\mathcal R$.

Provenance overlap and spatial concentration are not equivalent: two evidence maps can have the same overlap with $U$ or $P_Q$ but distribute their remaining mass differently. For sign $s\in\{+,-\}$, let $p_{\ell,s}(v)=E^s_{\ell,\hat y}(v)/ \max(\sum_u E^s_{\ell,\hat y}(u),\epsilon)$ and define normalized entropy concentration
\begin{equation}
h_{\ell,s}=1-
\frac{-\sum_v p_{\ell,s}(v)\log(p_{\ell,s}(v)+\epsilon)}
{\log |V|}.
\label{eq:concentration}
\end{equation}
The concentration-family risks are $r^H_{\ell,+}=\kappa_{\hat y}h_{\ell,+}$ and $r^H_{\ell,-}=-\kappa_{\hat y}h_{\ell,-}$: narrowly concentrated supportive evidence and diffusely distributed contradiction are treated as risk candidates. These are diagnostic hypotheses rather than grounding claims: localized support can reveal reliance on one visual shortcut, whereas image-wide contradiction can reveal broadly conflicting evidence. Inner grouped validation retains this family only when those patterns predict errors within calibration. This yields $\mathcal R^{H}=\{r^H_{\ell,+},r^H_{\ell,-}\}_{\ell}$. A route is one scalar indexed by family, layer, and sign.

\subsection{Top-$k$ Portfolio and Fusion}

Candidate routes can be noisy if used uniformly, so WEP treats them as a small portfolio. For each outer held-out fold, the calibration images are first split into three inner grouped folds. Inner out-of-fold AP selects $F^\star\in\{P,H\}$, choosing between provenance and concentration without access to outer-fold labels. Using all outer-calibration images, routes in $\mathcal R^{F^\star}$ are ranked by individual error AP; portfolio size $k$ and fusion weight $\beta$ are chosen over predefined grids. These choices are then frozen for the held-out images. Selected routes are standardized with calibration-fold statistics and averaged:
\begin{equation}
r_{\mathrm{ev}}(x)=
\frac{1}{k}\sum_{m\in\mathcal{P}_k}
\frac{r_m(x)-\mu_m}{\sigma_m}.
\label{eq:portfolio}
\end{equation}
The final score fuses evidence with confidence:
\begin{equation}
r_{\mathrm{wep}}(x)=
\frac{r_{\mathrm{conf}}(x)-\mu_c}{\sigma_c}
\;+\;
\beta
\frac{r_{\mathrm{ev}}(x)-\mu_e}{\sigma_e}.
\label{eq:fusion}
\end{equation}
The family, route set, normalization statistics, and $\beta$ are fixed after validation. No per-route weights are learned; the controller adds one binary family choice to the original sparse portfolio. The portfolio is what lets the same carrots-style failure be detected through the few layer/sign routes that are stable for the current backbone and benchmark, rather than through a single hand-chosen layer.

\subsection{Complexity}

The detector uses quantities from the instrumented prefill pass plus sparse row recomputation when attention weights are not materialized. Efficient-attention models can recompute only question-token rows for $P_Q$ and the final decision row for $C_{\ell,y}$ from cached $Q,K$, avoiding the full attention matrix. If $N_v$ is the number of visual tokens and $|Q|$ the number of question tokens, the row-level attention work scales as $O(LH(|Q|+1)N_v)$ rather than $O(LHN^2)$. The value readout adds only candidate-margin dot products, and the stored report contains logits, metadata, confidence, and compact route scalars.

WEP is therefore a scoring layer over one instrumented forward path, usable offline for benchmark auditing or online before accepting, abstaining, or routing an answer. The supplementary latency audit reports the overhead of our current Python-hook implementation separately from this path-count claim.

\section{Experiments}

\begin{table*}[t]
\centering
\small
\setlength{\tabcolsep}{2pt}
\begin{tabular}{@{}p{0.13\textwidth}p{0.12\textwidth}rrrp{0.18\textwidth}rrp{0.18\textwidth}@{}}
\toprule
& & & \multicolumn{3}{c}{AP} & \multicolumn{3}{c}{AUROC} \\
\cmidrule(lr){4-6}\cmidrule(lr){7-9}
Model & Dataset & Err. & Conf. & WEP & $\Delta$ [95\% CI] & Conf. & WEP & $\Delta$ [95\% CI] \\
\midrule
Qwen3-VL-8B & AMBER-D & 0.143 & 0.334 & 0.438 & \textbf{+0.104 [+0.087, +0.118]} & 0.741 & 0.806 & \textbf{+0.065 [+0.057, +0.072]} \\
 & Causal-HalBench & 0.167 & 0.374 & 0.468 & \textbf{+0.094 [+0.076, +0.109]} & 0.773 & 0.789 & \textbf{+0.016 [+0.011, +0.022]} \\
 & VSR & 0.202 & 0.373 & 0.411 & +0.038 [-0.001, +0.074] & 0.740 & 0.738 & -0.002 [-0.015, +0.010] \\
 & HallusionBench & 0.417 & 0.458 & 0.508 & \textbf{+0.050 [+0.009, +0.089]} & 0.527 & 0.605 & \textbf{+0.078 [+0.030, +0.122]} \\
\addlinespace[2pt]
LLaVA-1.5-7B & AMBER-D & 0.257 & 0.408 & 0.611 & \textbf{+0.203 [+0.187, +0.216]} & 0.689 & 0.812 & \textbf{+0.123 [+0.116, +0.130]} \\
 & Causal-HalBench & 0.187 & 0.420 & 0.588 & \textbf{+0.168 [+0.148, +0.190]} & 0.775 & 0.846 & \textbf{+0.071 [+0.063, +0.080]} \\
 & VSR & 0.308 & 0.472 & 0.489 & +0.017 [-0.022, +0.054] & 0.686 & 0.692 & +0.006 [-0.009, +0.021] \\
 & HallusionBench & 0.501 & 0.526 & 0.612 & \textbf{+0.086 [+0.042, +0.120]} & 0.509 & 0.605 & \textbf{+0.096 [+0.054, +0.138]} \\
\addlinespace[2pt]
InternVL3.5-8B & AMBER-D & 0.279 & 0.468 & 0.822 & \textbf{+0.354 [+0.339, +0.370]} & 0.711 & 0.909 & \textbf{+0.198 [+0.191, +0.205]} \\
 & Causal-HalBench & 0.452 & 0.617 & 0.887 & \textbf{+0.270 [+0.252, +0.286]} & 0.756 & 0.921 & \textbf{+0.165 [+0.156, +0.174]} \\
 & VSR & 0.387 & 0.600 & 0.761 & \textbf{+0.161 [+0.123, +0.199]} & 0.767 & 0.828 & \textbf{+0.060 [+0.040, +0.080]} \\
 & HallusionBench & 0.479 & 0.567 & 0.631 & \textbf{+0.064 [+0.026, +0.107]} & 0.585 & 0.642 & \textbf{+0.057 [+0.026, +0.092]} \\
\midrule
Mean & 12 pairs & -- & 0.468 & 0.602 & \textbf{+0.134} & 0.688 & 0.766 & \textbf{+0.078} \\
\bottomrule
\end{tabular}
\caption{Image-grouped out-of-fold error detection. WEP selects the provenance or signed-concentration route family by inner grouped validation, then selects and fuses routes using only the outer training fold. CIs are paired image-cluster bootstrap intervals for deltas over confidence.}
\label{tab:main}
\end{table*}

\begin{figure*}[!t]
\centering
\includegraphics[width=0.62\textwidth]{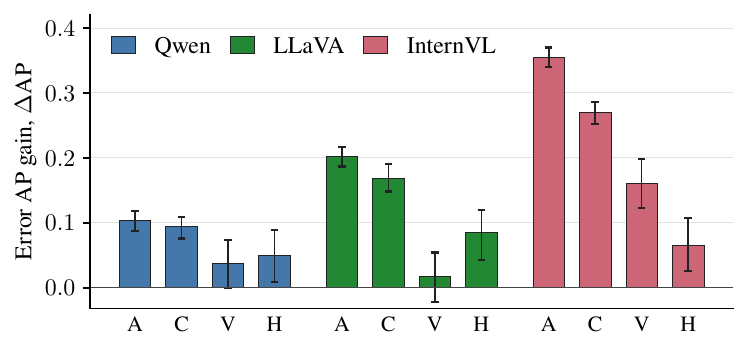}
\hfill
\includegraphics[width=0.34\textwidth]{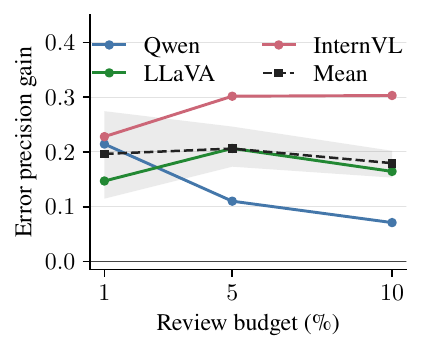}
\caption{Error-ranking and review-budget gains. Left: AP gains over confidence with 95\% image-cluster bootstrap intervals. Right: mean review-budget precision gain by model and review fraction. WEP is strongest near the top of the risk list, the common operating point for audit or selective routing. Within each model block, A/C/V/H denote AMBER-D, Causal-HalBench, VSR, and HallusionBench.}
\label{fig:analysis}
\end{figure*}

\noindent\textbf{Research questions.} RQ1 asks whether WEP ranks errors better than confidence; RQ2 tests which evidence components produce the gain and whether they remain matched to the current example and question; RQ3 evaluates calibration, stability, and transfer.

\subsection{Experimental Setup}

\noindent\textbf{Models and benchmarks.} We evaluate three open MLLMs, Qwen3-VL-8B \citep{bai2025qwen3vl}, LLaVA-1.5-7B \citep{liu2024llava15,liu2023llava}, and InternVL3.5-8B \citep{chen2025internvl35}, on four closed-answer visual reasoning and hallucination benchmarks: AMBER-D, Causal-HalBench, VSR, and HallusionBench \citep{wang2023amber,xu2025causalhalbench,liu2023vsr,guan2024hallusionbench}. Together they cover object, attribute, relation, counterfactual, spatial, and illusion/language-prior errors. The scored probe counts per backbone are 14,216 for AMBER-D, 9,709 for Causal-HalBench, 1,249 for VSR, and 674 for HallusionBench. All four main benchmarks use the single-token candidates \texttt{yes/no}; four-choice and multi-token finite-candidate prototypes are reported in the supplement. Question-only entity spans are recovered for 1,196/1,249 VSR questions; the remaining VSR items and the other three benchmarks use the full-question map $P_Q$.

\noindent\textbf{Metrics and baselines.} Our primary metric is error average precision (AP), which ranks incorrect current predictions above correct ones and emphasizes the high-risk tail used in audit or routing. We also report AUROC, review-budget precision, and 1000-replicate image-cluster bootstrap intervals. The main baseline is negative candidate margin, the natural single-prefill confidence signal. We further compare provenance-only and concentration-only portfolios, internal-evidence ablations, L1 and stability-selected route residuals, a high-capacity in-domain all-route logistic reference, and VCD/ICD extra-path controls that contrast margins after image or instruction perturbation. The supplementary protocol table separates inference paths, white-box access, labeled calibration, and capacity. PAS, VADE, and Attention Never Lie predict generated objects, tokens, or windows from attention patterns, so a direct adaptation would change both the prediction unit and supervision. We therefore compare protocol-matched single-prefill controls: candidate-adapted Overthinking and trajectory scores, HALP-QT at its source-supported $3L/4$ depth, and learned selectors over all-layer decision or pooled multimodal states. These matched-budget analogues of selective-VQA selectors \citep{kogilathota2026halp,whitehead2022reliablevqa,dancette2023selectivevqa} use the same predictions, labels, folds, and calibration examples as WEP but no witness or route features.

The dense all-provenance-route logistic reference is strongest in-domain ($+0.163$ mean $\Delta$AP), but learns a coefficient for every route. WEP trades $0.029$ mean $\Delta$AP for a named, unweighted sparse portfolio and one fusion weight; RQ3 tests whether this lower-capacity structure transfers more reliably without target labels.

\noindent\textbf{Protocol.} WEP uses one instrumented prefill. Within each outer calibration split, three image-grouped inner folds select the route family; the full split then selects route identities, $k\in\{1,2,3,5,8,13,21,36\}$, and one fusion weight. These choices, which include no per-route coefficients, are frozen on the held-out images. Intervals resample image clusters, while a separate 50-seed analysis reruns fold construction and selection. We also test a predefined provenance family when target labels cannot select a family, plus evidence-shuffle nulls. The supplement gives exact prompt construction, candidate scoring, and implementation details.

\subsection{RQ1: Error Ranking and Triage}

WEP consistently improves high-risk error ranking across models and datasets. In the evaluated closed-answer setting, Table~\ref{tab:main} gives the full 12-pair result: WEP has positive AP gains on every pair, with mean $\Delta$AP $+0.134$; 10 of 12 image-cluster bootstrap intervals are strictly positive. Backbone-wise mean gains are $+0.071$ for Qwen, $+0.119$ for LLaVA, and $+0.212$ for InternVL, so the result is not solely an InternVL effect despite its larger absolute gains. AUROC and per-dataset variation are reported in the table.

Figure~\ref{fig:analysis} shows operationally meaningful triage gains. On average, precision improves by $+0.196$ [$+0.115,+0.274$], $+0.206$ [$+0.173,+0.246$], and $+0.179$ [$+0.153,+0.202$] at 1\%, 5\%, and 10\% review budgets, respectively; intervals use image-cluster resampling. All three backbone means remain positive at each budget. The supplement reports AURC, risk--coverage, and fixed-review routing metrics.

\subsection{RQ2: Evidence Components}

\noindent\textbf{Formula components.} Tables~\ref{tab:ablation} and~\ref{tab:extra-path-main} isolate the main components. Unsigned evidence gives only $+0.014$ mean $\Delta$AP and uniform route averaging $+0.081$. Provenance and concentration alone reach $+0.129$ and $+0.124$ but win on different cells; nested selection reaches $+0.134$, choosing them in 37 and 23 of 60 outer folds. Replacing the provenance question residual with a matched other question lowers AP by $0.009$ [$0.002,0.016$]. At seed 0, residuals occupy 300/617 selected provenance slots and 38.9\% of standardized absolute pre-fusion route mass. Pooling both families under the same top-$k$ capacity reaches $+0.128$, below nested selection, so an enlarged joint route pool is unnecessary. Removing question-residual routes gives $+0.127$; full WEP is higher on 9/12 cells, showing a modest complementary gain. This removes only explicit witness-minus-uniform alignment: retained routes remain conditioned on the current image--question--candidate path through $E$.

\noindent\textbf{Extra-path baselines.} On the strict same-prediction subset, WEP has mean $\Delta$AP $+0.127$, while fixed VCD-margin, ICD-margin, and the best diagnostic VCD sweep give $-0.101$, $-0.167$, and $-0.023$. These controls use additional image or instruction paths, whereas WEP scores the original answer from the same prefill used to obtain confidence.

\begin{table}[t]
\centering
\footnotesize
\setlength{\tabcolsep}{2pt}
\begin{tabular}{@{}p{0.45\columnwidth}p{0.20\columnwidth}rr@{}}
\toprule
Variant & Test & Mean $\Delta$AP & Positive \\
\midrule
Confidence only & baseline & +0.000 & 0/12 \\ 
Unsigned evidence & sign & +0.014 & 8/12 \\ 
No visual transport & transport & +0.054 & 2/3 \\ 
Uniform route mean & selection & +0.081 & 11/12 \\ 
Provenance only & family & +0.129 & 12/12 \\
Concentration only & family & +0.124 & 12/12 \\
Nested WEP & family select. & \textbf{+0.134} & 12/12 \\ 
\bottomrule
\end{tabular}
\caption{Formula-level ablations. The last three rows use image-grouped outer folds; nested WEP selects a family by inner grouped validation. No-transport is evaluated on the three-backbone VSR subset.}
\label{tab:ablation}
\end{table}

\begin{table}[t]
\centering
\small
\setlength{\tabcolsep}{3pt}
\begin{tabular}{p{0.38\columnwidth}ccc}
\toprule
Risk score & Extra path & Mean $\Delta$AP & AP wins \\
\midrule
Confidence margin & No & +0.000 & 1/9 \\
VCD-margin & Image & -0.101 & 0/9 \\
ICD-margin & Prompt & -0.167 & 0/9 \\
Best VCD sweep & Image & -0.023 & 0/9 \\
WEP & No & +0.127 & 8/9 \\
\bottomrule
\end{tabular}
\caption{Extra-path contrastive controls on the strict same-prediction subset.
Mean $\Delta$AP is relative to the confidence margin within each cell.}
\label{tab:extra-path-main}
\end{table}

\noindent\textbf{Matched-budget internal baselines.} These controls use deterministic same-row subsets and, for zero-target transfer, fix the provenance family so no target labels choose between route families. Candidate-adapted Overthinking and its trajectory LR reach $0.334$ and $0.528$ AP, respectively. HALP-QT \citep{kogilathota2026halp} reaches $0.610\pm0.003$ versus $0.592$ for the provenance fallback ($-0.018$ [$-0.036,+0.001$]); their respective AUROC and AURC values are $0.759$/$0.187$ and $0.762$/$0.184$. A projected-state MLP reaches $0.594\pm0.004$. In leave-dataset-out transfer, provenance gives $+0.089$ AP over confidence, versus $-0.065$ for HALP-QT and $-0.058$ for the MLP.

\noindent\textbf{Evidence matching and local relevance.} Permuting evidence across examples while fixing confidence and labels turns the matched provenance gain negative on average, tying improvement to matched internal evidence rather than a score-distribution shift. Full-cell score and route-reselection label nulls support the same conclusion.

Complementary mechanism checks connect the local estimator to final decisions. Value patching intervenes at the scored layer, executes every remaining transformer block, and measures the final candidate margin. Selected-versus-control gaps are positive on VSR and HallusionBench, as is Causal-HalBench support-versus-contradiction patching. Across 64 AMBER-D examples per backbone, the local source readout has $0.018$--$0.082$\% median closure error and $0.975$--$0.994$ rank correlation with exact same-sublayer margin changes. Final-margin effects are nonzero in 13/18 fixed backbone--dataset--depth tests; their depth-dependent signs motivate layer-specific route selection rather than uniform aggregation. Grounded-VSR witness maps beat matched references on all three. Oracle geometry raises VSR macro AP from $0.540$ to $0.562$, locating remaining spatial errors beyond route selection. Full audits and maps are supplementary.

\subsection{RQ3: Calibration, Stability, and Transfer}

\noindent\textbf{Calibration and stability.} With 100/250 labels, the predefined provenance family gives mean $\Delta$AP $+0.069/+0.107$. Across 50 grouped-fold seeds, complete nested reselection of family, routes, $k$, normalization, and fusion gives $+0.129$ [$+0.126,+0.134$], with 581/600 cell--seed gains positive. A separate 50-trial provenance-label null is near zero and has empirical $p\leq0.039$ in 11/12 cells. All 60 final outer-fold fusion coefficients are positive.

\noindent\textbf{Capacity and transfer fallback.} Full-label and stability L1 reach $0.641/0.601$ AP with 80.2/28.5 weighted routes; provenance reaches $0.593$ with 15.0 and no per-route coefficients. Without target labels for family selection, provenance transfers at $+0.095$ [$+0.085,+0.102$] on 11/12 targets versus $+0.073/+0.070$ on 10/12 for the two sparse learners. Evaluation manifests have no cross-dataset image-hash overlap; latency audits are supplementary.

\section{Discussion}

\noindent\textbf{Interpretation.} Confidence measures candidate separation; WEP asks what visual evidence accompanies it. Provenance tests whether evidence follows question-bound witnesses, while concentration tests whether signed support is compact or diffuse. Their different fold-level selections explain why one groundedness scalar is insufficient, and the shuffle and patching controls connect gains to example-matched readouts rather than score-scale artifacts.

\noindent\textbf{Operational use.} WEP leaves the answer unchanged and raises weakly supported or contradictory predictions in a review queue. Deployments can set review or routing thresholds to available capacity \citep{angelopoulos2021conformal,wang2026lec}, while retaining named route contributions for audit. WEP supplies the ranking signal; the policy determines which predictions to accept, review, or reroute.

\noindent\textbf{Scope.} WEP assumes white-box readouts, finite candidates, and labeled calibration. The main study is binary; free-form use requires claim segmentation and aggregation. Spatial reasoning leaves more geometric headroom than existence or attribute questions, and sparse read-row recomputation still requires serving-kernel integration.

\section{Conclusion}

We introduced WEP, a single-prefill detector that augments candidate confidence with signed visual-evidence provenance and concentration. Across three MLLMs and four benchmarks, its sparse portfolios consistently improve error ranking and high-risk review. Shuffle, transfer, localization, and intervention analyses link the gains to internal decision evidence. The results establish question-related visual provenance as a practical signal for deciding which closed multimodal answers warrant review or routing.

\bibliography{qwdc}

\appendix \raggedbottom
\section{Supplementary Material}

\section{Protocol Details}

\textbf{Task format.} All main experiments use the closed candidate set \{\texttt{yes}, \texttt{no}\}. The formulation permits larger finite sets, and a four-choice MMStar pilot and a two-alternative, multi-token prototype are reported in Tables~\ref{tab:mmstar-pilot} and~\ref{tab:wep-span}. The detector ranks the model's current candidate-set prediction by error risk and does not alter that prediction. The main prediction is, by definition, the constrained candidate-set argmax; agreement with unconstrained greedy generation is not an inclusion criterion. The four-choice pilot reports this agreement separately.

\textbf{Closed-answer probe construction.} We convert the public benchmark releases without answer balancing or manual example selection. For AMBER-D, release annotations are joined to discriminative queries by sample id; we retain every row with a nonempty question, a resolvable image, and an answer normalizable to \texttt{yes}/\texttt{no}. For Causal-HalBench, we retain every released row meeting the same three conditions, without filtering its tag or counterfactual type. Boolean and numeric labels are mapped deterministically to \texttt{yes}/\texttt{no}, literal \texttt{<image>} markers are removed, and the original category or tag/type is preserved as metadata. We do not collapse questions sharing an image: the resulting 14,216 AMBER-D probes cover 1,004 images, and 9,709 Causal-HalBench probes cover 2,144 images; image grouping keeps all such questions in one fold. The submitted JSONL files contain the exact resulting probe rows and relative image indices.

\textbf{Nested fold selection.} For each model--dataset pair, we split images into five outer folds with a fixed seed; all questions sharing an image remain in the same fold. For each outer fold, three inner image-grouped folds compare provenance and concentration by out-of-fold error AP. The winning family is then refit on all outer-calibration images: route ranking, normalization, portfolio size, and fusion weight use no held-out images. The outer fold is scored with these frozen choices, producing the main-paper results. The reported fold seed is 0. The complete stability audit reruns seeds 0--49, and the main image-cluster bootstrap uses a deterministic RNG initialized from seed 2027. Model scoring uses evaluation/inference mode without sampling.

\textbf{Span source.} Entity strings are parsed from the visible question. A question-only audit with no caption, relation field, box, target, or benchmark annotation recovers all 1,196 pair-specific VSR parses exactly and reproduces all 53 full-question fallbacks used by the cached experiment. Tokenizer offsets locate the recovered phrases inside the question. Both entity maps fall back to the full-question witness when either phrase is missing. This yields pair-specific maps for 1,196/1,249 VSR items (95.8\%); the remaining 53 VSR items and all AMBER-D, Causal-HalBench, and HallusionBench items use the full-question witness because those probes do not supply a canonical subject--object pair. For these fallback cases $S=O=P_Q$, so the support minimum and contradiction maximum reduce to the same full-question overlap; the two-entity rule is active only for examples with paired spans. For VSR, the deterministic parser strips terminal punctuation, rewrites \texttt{Does X contain Y} as \texttt{X contains Y}, otherwise removes an initial \texttt{Is}/\texttt{Are}, and matches the longest phrase in a fixed spatial-relation lexicon. Entity token spans are then located by tokenizer offsets, with token-subsequence matching as a fallback. No answer, image annotation, or label enters this procedure.

\textbf{Low-capacity controller.} The controller adds one discrete family choice, a small route set, and one fusion weight, but learns no per-route coefficients. After selection, route risks are standardized with training-fold statistics, averaged, and fused with standardized confidence risk. Dense linear and nonlinear selectors are therefore high-capacity references rather than the deployment method.

\textbf{Transfer fallback.} Zero-target leave-dataset-out transfer excludes the target from route selection, standardization, and fusion. Because target labels cannot choose a family, we fix provenance, select its portfolio on the other three datasets of the same backbone, and apply it unchanged. Projected and full-state selectors use the same backbone-matched training scope; the stricter linear stack trains on all other model--dataset groups.

\textbf{Sanity controls.} The evidence-shuffle control keeps each example's confidence score and label fixed but permutes evidence scores across examples. It preserves marginal score distributions while destroying example-specific alignment; its near-zero or negative gain rules out a generic score-scale explanation.

Table~\ref{tab:method-terms} summarizes the notation used by the operational definitions below.
\begin{table}[tbp]
\centering
\footnotesize
\setlength{\tabcolsep}{2pt}
\begin{tabular}{@{}p{0.16\columnwidth}p{0.28\columnwidth}p{0.46\columnwidth}@{}}
\toprule
Level & Object & Role \\
\midrule
Region & $P_Q,S,O,U$ & question witnesses and uniform reference \\
Source & $C_{\ell,y}(s)$ & candidate-margin contribution \\
Evidence & $E_{\ell,y}(v)$ & signed contribution in image space \\
Route family & $\mathcal R^P,\mathcal R^H$ & provenance or concentration view \\
Route & $r^G_{\ell,\pm},r^\Delta_{\ell,\pm},r^H_{\ell,\pm}$ & one family/layer/sign risk scalar \\
Portfolio & $r_{\mathrm{ev}}$ & mean risk of selected routes \\
Final score & $r_{\mathrm{wep}}$ & evidence plus confidence risk \\
\bottomrule
\end{tabular}
\caption{WEP terminology. Inner validation selects one route family; the
portfolio then averages a validation-selected set of its layer/sign routes.}
\label{tab:method-terms}
\end{table}

\section{Implementation and Search Details}

\textbf{Checkpoints and image processing.} We use the public checkpoints \texttt{Qwen/Qwen3-VL-8B-Instruct}, \texttt{llava-hf/llava-1.5-7b-hf}, and \texttt{OpenGVLab/InternVL3\_5-8B-HF} with their released processors and tokenizers. Qwen fixes both minimum and maximum image area to 50,176 pixels with patch size 16 and spatial merge size 2. LLaVA resizes and center-crops to $336\times336$ with patch size 14. InternVL uses $448\times448$ tiles, dynamic tiling with 1--12 patches, and its native thumbnail behavior. Qwen and InternVL use SDPA; LLaVA uses eager attention so its sparse rows can be recomputed exactly. All models run in evaluation mode with their released mixed-precision dtype.

\textbf{Prompt construction.} For the four main binary benchmarks, we pass each canonical question unchanged after removing a literal \texttt{<image>} placeholder; no candidate list or extra yes/no instruction is appended. Qwen uses its native chat template with the system message \texttt{You are a helpful assistant.}; InternVL uses its native chat template; and LLaVA uses the processor template, with the standard Vicuna-style template only as a fallback. Candidate logits are read at the assistant generation position of these templates.

\textbf{Candidate scoring.} All candidates are scored from the same prefill state. We tokenize each candidate with the backend's evaluation delimiter: a leading space for Qwen and InternVL, and no added delimiter for LLaVA. All evaluated yes/no labels map to one token under these prompts. The confidence risk is the negative logit margin between the top two candidates. WEP never teacher-forces alternative answers and does not alter the candidate-set argmax.

\textbf{Hooks and stored tensors.} For Qwen3-VL, LLaVA-1.5, and InternVL, the hooks are placed on the language transformer blocks after visual tokens have been inserted into the multimodal sequence. During the original prefill, we keep candidate logits, visual-token indices, the question-token rows needed for witness maps, and per-layer candidate-margin contribution summaries. Full visual token maps are stored only for localization and patching audits; the main risk reports store compact route scalars rather than full token maps.

\textbf{Route enumeration.} For every stored layer $\ell$, WEP enumerates two signs within two candidate families. The provenance family contains the global-reference and question-residual bases, $\mathcal R^P=\{r^G_{\ell,\pm},r^\Delta_{\ell,\pm}\}_{\ell}$. The concentration family contains normalized-entropy routes, $\mathcal R^H=\{r^H_{\ell,\pm}\}_{\ell}$. A route is one family/layer/sign scalar; no unavailable route is imputed.

\textbf{Exact operational definitions.} For a nonnegative vector $a$, let $\mathcal{N}_{+}(a)=a/\max(\sum_i a_i,\epsilon)$, preserving an all-zero vector as zero. Let
\[
 M_{\ell}(a,b)=\sum_h A_{\ell,h}(a,b)
 \lVert V_{\ell,h}(b)\rVert_2
 \frac{\lVert W_O^{\ell,h}\rVert_F}{\sqrt{d_h}}
\]
denote the attention/value read mass used by the hooks. The implementation normalizes each question-to-image read row and averages it over stored layers:
\[
 B_j=\mathcal{N}_{+}\!\left(
 \frac{1}{L}\sum_{\ell=1}^{L}
 \mathcal{N}_{+}(M_{\ell}(j,V))\right).
\]
The question-token weight is $\omega_j=\mathcal{N}_{+}(\sum_\ell M_\ell(t,j))$, where $t$ is the decision position. A question token transports only its visually sourced fraction
\[
 \rho_j=
 \frac{\sum_{\ell,v\in V}M_\ell(j,v)}
      {\max(\sum_{\ell,s}M_\ell(j,s),\epsilon)}\in[0,1],
\]
so the exact transported evidence is
\[
\begin{aligned}
 \bar C_{\ell,y}(s)&=
 \frac{C_{\ell,y}(s)}{\max(\sum_u|C_{\ell,y}(u)|,\epsilon)},\\
 E_{\ell,y}(v)&=\bar C_{\ell,y}(v)+
 \sum_{j\in Q}\bar C_{\ell,y}(j)\rho_jB_j(v).
\end{aligned}
\]
The remaining text-sourced mass is assigned to a null bucket and does not enter witness overlap.

For an entity span $\mathcal{E}$, its raw map is the normalized mean of $B_j$ over $j\in\mathcal{E}$. Control tokens are all visible question tokens outside the parsed subject, object, and relation spans; $H$ is their normalized mean binding map (zero when this set is empty). With $\lambda_{\rm sink}=1.0$, the entity witness is
\[
 W_{\mathcal{E}}=\mathcal{N}_{+}\!\left([\operatorname{raw}(\mathcal{E})-
 \lambda_{\rm sink}H]_+\right);
\]
if subtraction removes all mass, the raw map is retained. Subject and object maps use $S=W_{\rm subj}$ and $O=W_{\rm obj}$; unavailable spans use $S=O=P_Q$. Before overlap, evidence is attenuated by $\max(0.05,1-H(v)/\max_u H(u))$ and normalized by its visual $\ell_1$ mass. The global reference is $U(v)=1/|V|$ over valid visual tokens.

Finally, let $m_{\ell,y}$ be the candidate margin read from layer $\ell$. The layer gate and candidate-level visually sourced coverage are
\[
\begin{aligned}
 \gamma_{\ell,y}&=
 \frac{|m_{\ell+1,y}-m_{\ell,y}|}
      {\max(\sum_r|m_{r+1,y}-m_{r,y}|,\epsilon)},\\
 \kappa_y&=\sum_\ell\gamma_{\ell,y}
 \sum_s|\bar C_{\ell,y}(s)|\\
 &\quad\cdot\sum_{v\in V}\Pi_s(v).
\end{aligned}
\]
where $\Pi_s$ is the source-to-visual transport distribution. The overlap factor is $\kappa_y$, shared across layers; $\gamma_{\ell,y}\kappa_y$ is evaluated only as a per-layer-gated ablation. Positive and negative evidence are separately normalized before histogram-intersection overlap. Let $r^Q_{\ell,+}=-g^{Q,+}_{\ell}$ and $r^Q_{\ell,-}=g^{Q,-}_{\ell}$ denote risks computed with $(S,O)$. Replacing both maps by $U$ gives $r^G_{\ell,\pm}$. The final question-specific route is
\[
 r^\Delta_{\ell,\pm}=r^Q_{\ell,\pm}-r^G_{\ell,\pm}.
\]
Although the distribution $U$ is question-independent, $r^G$ remains conditioned on the current image--question--candidate path through $E_{\ell,y}$; ``global'' denotes its uniform overlap reference, not an image-only computation. Portfolio selection receives only $\{r^G_{\ell,\pm},r^\Delta_{\ell,\pm}\}_\ell$; the raw question route is not an additional provenance feature. For the concentration family, separately normalize $E^+_{\ell,y}$ and $E^-_{\ell,y}$ over visual tokens and define
\[
h_{\ell,s}=1-
\frac{-\sum_v p_{\ell,s}(v)\log(p_{\ell,s}(v)+\epsilon)}
{\log |V|}.
\]
The risk-oriented routes are $r^H_{\ell,+}=\kappa_y h_{\ell,+}$ and $r^H_{\ell,-}=-\kappa_y h_{\ell,-}$. These definitions are shared by the three backbones; only the hook exposing $A,V,W_O$, and candidate logits differs.

\textbf{Nested family and portfolio search.} Three inner grouped folds independently fit a provenance portfolio and a concentration portfolio, and their pooled inner out-of-fold AP chooses the family. On all outer-calibration images, individual routes from the selected family are ranked by error AP. The candidate portfolio sizes are \(\{1,2,3,5,8,13,21,36\}\), clipped to the number of available routes. For each candidate size, the selected routes are standardized using training-fold statistics, averaged, and fused with standardized confidence risk. The fusion grid is \(\beta\in\{-3.00,-2.75,\ldots,2.75,3.00\}\). The held-out fold receives only the chosen route identities, training-fold statistics, portfolio size, family identity, and \(\beta\), and is excluded from selection.

\textbf{Determinism and tie-breaking.} All fold splits use fixed seeds. Route ranking and configuration names are sorted deterministically before selection. If two portfolios have the same selection score up to floating-point equality, the implementation breaks ties with the recorded AP, AUROC, portfolio size, fusion weight, and route-name ordering, making the selected portfolio reproducible under the reported tie-breaking rule.

\textbf{Bootstrap intervals.} Paired bootstrap intervals resample complete image clusters from the frozen out-of-fold confidence and WEP scores; route selection is not rerun inside each resample. The intervals therefore quantify resampling uncertainty conditional on the realized out-of-fold selections. The separate 50-seed analysis reruns the complete final nested algorithm---outer folds, inner family comparison, route identities, portfolio size, normalization, and fusion---and therefore measures algorithmic selection sensitivity. We report these two distinct uncertainty sources separately rather than interpreting the conditional bootstrap as uncertainty over the learning procedure.

\textbf{Candidate-token coverage.} After closed-answer probe construction, all four main benchmark evaluations use single-token yes/no candidate labels under the prompts used for the three backbones. Thus no additional examples are dropped by model tokenizer at candidate scoring time; the scored probe counts are the $n$ values in Table~\ref{tab:app-full-main}. These counts should not be read as the raw sizes of the upstream benchmark releases.

\section{Expanded Main Results}

Table~\ref{tab:app-full-main} expands the compact main-paper summary with per-cell review precision alongside the final nested AP results.

\begin{table*}[!tbp]
\centering
\small
\setlength{\tabcolsep}{2.4pt}
\begin{tabular}{@{}llrrrrrrrrr@{}}
\toprule
Model & Data & $n$ & Err. & Conf. AP & WEP AP & $\Delta$AP & \multicolumn{2}{c}{P@5\%} & \multicolumn{2}{c}{P@10\%} \\
\cmidrule(lr){8-9}\cmidrule(lr){10-11}
 & & & & & & & Conf. & WEP & Conf. & WEP \\
\midrule
Qwen & AMB. & 14216 & 0.143 & 0.334 & 0.438 & +0.104 & 0.459 & 0.579 & 0.424 & 0.499 \\
Qwen & Causal & 9709 & 0.167 & 0.374 & 0.468 & +0.094 & 0.477 & 0.638 & 0.443 & 0.546 \\
Qwen & VSR & 1249 & 0.202 & 0.373 & 0.411 & +0.038 & 0.429 & 0.587 & 0.456 & 0.488 \\
Qwen & Hall. & 674 & 0.417 & 0.458 & 0.508 & +0.050 & 0.559 & 0.559 & 0.500 & 0.574 \\
LLaVA & AMB. & 14216 & 0.257 & 0.408 & 0.611 & +0.203 & 0.473 & 0.809 & 0.458 & 0.731 \\
LLaVA & Causal & 9709 & 0.187 & 0.420 & 0.588 & +0.168 & 0.484 & 0.774 & 0.484 & 0.702 \\
LLaVA & VSR & 1249 & 0.308 & 0.472 & 0.489 & +0.017 & 0.524 & 0.635 & 0.536 & 0.584 \\
LLaVA & Hall. & 674 & 0.501 & 0.526 & 0.612 & +0.086 & 0.647 & 0.735 & 0.529 & 0.647 \\
InternVL & AMB. & 14216 & 0.279 & 0.468 & 0.822 & +0.354 & 0.534 & 0.931 & 0.521 & 0.916 \\
InternVL & Causal & 9709 & 0.452 & 0.617 & 0.887 & +0.270 & 0.572 & 0.942 & 0.571 & 0.931 \\
InternVL & VSR & 1249 & 0.387 & 0.600 & 0.761 & +0.161 & 0.571 & 0.952 & 0.592 & 0.872 \\
InternVL & Hall. & 674 & 0.479 & 0.567 & 0.631 & +0.064 & 0.706 & 0.765 & 0.618 & 0.794 \\
\bottomrule
\end{tabular}
\caption{Expanded final results. P@5\% and P@10\% are error precision among the corresponding highest-risk review queues.}
\label{tab:app-full-main}
\end{table*}

\begin{table}[H]
\centering
\small
\setlength{\tabcolsep}{3pt}
\begin{tabular}{lrrr}
\toprule
Metric & Confidence & WEP & $\Delta$ \\
\midrule
AURC $\downarrow$ & 0.214 & 0.179 & -0.035 \\
E-AURC $\downarrow$ & 0.147 & 0.112 & -0.035 \\
Acc.@90\% coverage $\uparrow$ & 0.707 & 0.727 & +0.020 \\
Error recall@5\% review $\uparrow$ & 0.096 & 0.135 & +0.039 \\
\bottomrule
\end{tabular}
\caption{Selective-prediction metrics averaged over the 12 model--dataset pairs using the final nested WEP scores.}
\label{tab:selective}
\end{table}

\textbf{Prevalence and review-budget robustness.} Because benchmark error prevalence varies substantially, Table~\ref{tab:statistical-robustness} recomputes the main comparison under prevalence-normalized, sample-weighted, and rank-pooled AP. All three views preserve a sizable WEP advantage. The same table reports image-cluster intervals for review precision; their lower endpoints remain positive at 1\%, 5\%, and 10\% budgets, showing that the gain is concentrated in the operational high-risk tail rather than being an artifact of one averaging convention.

\begin{table}[H]
\centering\small
\setlength{\tabcolsep}{4pt}
\begin{tabular}{lr}
\toprule
Robustness view & WEP gain \\
\midrule
Macro normalized AP & +0.201 \\
Sample-weighted AP & +0.193 \\
Rank-pooled AP & +0.187 \\
1\% review precision & +0.196 [+0.115,+0.274] \\
5\% review precision & +0.206 [+0.173,+0.246] \\
10\% review precision & +0.179 [+0.153,+0.202] \\
\bottomrule
\end{tabular}
\caption{Robustness to heterogeneous prevalence and review budget for final nested WEP. Review intervals use 1,000 image-cluster bootstrap replicates and macro-average the 12 cells.}
\label{tab:statistical-robustness}
\end{table}

\textbf{Formula-level and family ablations.} Table~\ref{tab:formula-ablation-full} expands the main-paper component study to individual model--dataset cells. The first block audits the provenance construction while retaining image-grouped out-of-fold selection and confidence fusion; the final rows compare provenance, concentration, and nested family selection. The no-transport row is restricted to the newly collected VSR records because older compact records did not store pre-transport source-token contributions.

\begin{table*}[!tbp]
\centering
\scriptsize
\setlength{\tabcolsep}{1.6pt}
\begin{tabular}{@{}lrrrrrrrrrrrrr@{}}
\toprule
& & \multicolumn{4}{c}{Qwen3-VL-8B} & \multicolumn{4}{c}{LLaVA-1.5-7B} & \multicolumn{4}{c}{InternVL3.5-8B} \\
\cmidrule(lr){3-6}\cmidrule(lr){7-10}\cmidrule(lr){11-14}
Variant & Mean & AMB. & Causal & VSR & Hall. & AMB. & Causal & VSR & Hall. & AMB. & Causal & VSR & Hall. \\
\midrule
Confidence only & +0.000 & +0.000 & +0.000 & +0.000 & +0.000 & +0.000 & +0.000 & +0.000 & +0.000 & +0.000 & +0.000 & +0.000 & +0.000 \\
Unsigned evidence & +0.014 & +0.010 & +0.020 & -0.008 & -0.008 & +0.005 & +0.040 & -0.026 & +0.050 & +0.029 & +0.068 & +0.009 & -0.026 \\
No visual transport & +0.054 & -- & -- & +0.024 & -- & -- & -- & -0.003 & -- & -- & -- & +0.140 & -- \\
No subject/object split & +0.110 & +0.041 & +0.023 & +0.021 & +0.040 & +0.135 & +0.186 & -0.003 & +0.068 & +0.353 & +0.278 & +0.139 & +0.045 \\
With per-layer margin gate & +0.101 & +0.024 & +0.060 & +0.013 & +0.020 & +0.158 & +0.127 & -0.001 & +0.095 & +0.344 & +0.248 & +0.105 & +0.021 \\
Question-only & +0.110 & +0.041 & +0.023 & +0.005 & +0.040 & +0.135 & +0.186 & +0.008 & +0.068 & +0.353 & +0.278 & +0.144 & +0.045 \\
Global-only & +0.121 & +0.063 & +0.081 & +0.027 & +0.027 & +0.184 & +0.188 & +0.008 & +0.071 & +0.343 & +0.263 & +0.138 & +0.059 \\
Residual-only & +0.108 & +0.084 & +0.082 & +0.030 & -0.007 & +0.089 & +0.150 & +0.002 & +0.058 & +0.361 & +0.266 & +0.176 & -0.001 \\
Global + raw & +0.122 & +0.064 & +0.062 & +0.071 & +0.025 & +0.162 & +0.186 & +0.011 & +0.070 & +0.351 & +0.278 & +0.139 & +0.051 \\
Uniform route mean & +0.081 & +0.087 & +0.070 & -0.007 & +0.003 & +0.058 & +0.085 & +0.000 & +0.033 & +0.316 & +0.242 & +0.061 & +0.027 \\
Global + witness + residual & +0.127 & +0.102 & +0.076 & +0.052 & +0.014 & +0.162 & +0.182 & +0.007 & +0.074 & +0.354 & +0.277 & +0.158 & +0.069 \\
\midrule
Provenance family & +0.129 & +0.104 & +0.086 & +0.008 & +0.048 & +0.203 & +0.168 & +0.017 & +0.086 & +0.354 & +0.270 & +0.149 & +0.056 \\
Concentration family & +0.124 & +0.064 & +0.097 & +0.055 & +0.055 & +0.168 & +0.170 & +0.015 & +0.060 & +0.330 & +0.261 & +0.148 & +0.064 \\
\textbf{Nested WEP} & \textbf{+0.134} & +0.104 & +0.094 & +0.038 & +0.050 & +0.203 & +0.168 & +0.017 & +0.086 & +0.354 & +0.270 & +0.161 & +0.064 \\
\bottomrule
\end{tabular}
\caption{Formula- and family-level ablations. Entries are image-grouped
out-of-fold AP gains over confidence. The upper block audits provenance
components; the lower block compares complete route families and nested
family selection. No-transport is available for the three-backbone VSR
records.}
\label{tab:formula-ablation-full}
\end{table*}

\textbf{Removing question-residual routes.} Table~\ref{tab:no-question-residual} restricts the provenance candidate to global routes while retaining its competition with the unchanged concentration family under the complete nested protocol. This controller retains $+0.127$ mean AP over confidence, versus $+0.134$ for full WEP; the full method is higher on 9/12 cells. Thus global signed evidence and concentration provide the main cross-task signal, while the residual supplies a smaller, question-matched increment rather than accounting for the gain by itself.

\begin{table}[H]
\centering
\footnotesize
\setlength{\tabcolsep}{3.5pt}
\begin{tabular}{@{}lrrr@{}}
\toprule
Controller & AP & $\Delta$ vs. conf. & Positive \\
\midrule
Confidence & 0.468 & -- & -- \\
Global/concentration nested & 0.595 & +0.127 & 12/12 \\
Full WEP & 0.602 & +0.134 & 12/12 \\
\bottomrule
\end{tabular}
\caption{Question-residual removal under the complete nested protocol. The
restricted controller selects between global signed-evidence and concentration
routes using the same grouped folds, top-$k$ grid, standardization, and fusion
grid as full WEP.}
\label{tab:no-question-residual}
\end{table}

\textbf{Matched-capacity joint route pool.} Table~\ref{tab:joint-union} removes the discrete family decision and places all provenance and concentration routes in one selectable pool. It keeps the same route-ranking rule, top-$k$ grid, standardization, and single fusion weight as WEP. The joint pool improves over confidence on all 12 cells but is $0.006$ AP below nested selection on average and exceeds it on only 2/12 cells. Thus family-first selection acts as a useful structural constraint rather than obtaining its gain from access to more fitted parameters.

\begin{table}[!tbp]
\centering
\footnotesize
\setlength{\tabcolsep}{3.0pt}
\begin{tabular}{lrrrrr}
\toprule
Model & Conf. & Nested & Joint & J$-$C & J$-$N \\
\midrule
Qwen & 0.385 & 0.456 & 0.451 & +0.066 & -0.006 \\
LLaVA & 0.457 & 0.575 & 0.569 & +0.113 & -0.006 \\
InternVL & 0.563 & 0.775 & 0.769 & +0.206 & -0.006 \\
\midrule
Macro & 0.468 & 0.602 & 0.596 & +0.128 & -0.006 \\
\bottomrule
\end{tabular}
\caption{Matched-capacity union control, averaged over each backbone's four
datasets. Joint uses the same top-$k$ grid and one fusion weight as nested WEP
but searches both route families in one pool. J$-$C and J$-$N denote joint
minus confidence and nested AP.}
\label{tab:joint-union}
\end{table}

\textbf{Image-fixed question swap within provenance.} Table~\ref{tab:question-swap} tests whether the selected evidence must match the current question rather than merely the image. For each eligible sample, we select the donor that shares the image and predicted candidate, uses another question, and has the nearest original candidate-confidence risk. This label-free matching controls for generic difficulty without using correctness labels. We replay the original fold configuration while keeping confidence, global routes, route ids, normalization, and fusion weight fixed. Swapping only residual routes lowers provenance-family AP by $0.009$ [$0.002,0.016$]; the residual-only detector drops by $0.027$ [$0.018,0.037$]. Intervals use a paired bootstrap over image clusters. The separate global-only control shows a larger drop, consistent with the global reference being the main anchor. This does not make the global route question-independent: both provenance bases inherit the current question's decision evidence, while only their overlap reference differs. Crucially, the intervention changes only the residual channel, showing that its smaller aggregate gain still contains information matched to the current question.

\begin{table}[H]
\centering
\footnotesize
\setlength{\tabcolsep}{3pt}
\begin{tabular}{@{}p{0.52\columnwidth}r@{}}
\toprule
Detector / swapped basis & AP drop [95\% CI] \\
\midrule
Provenance portfolio; swap residual & +0.009 [+0.002, +0.016] \\
Residual-only; swap residual & +0.027 [+0.018, +0.037] \\
Global-only; swap global & +0.082 [+0.068, +0.094] \\
\bottomrule
\end{tabular}
\caption{Image-fixed question-swap test within the provenance family. For each
eligible answer, route values are replaced by a donor sharing the image and
prediction but using another question with nearest confidence risk. Fold fits,
route ids, normalization, confidence, and fusion weights remain frozen.
Positive drops mean that question-matched routes rank errors better.}
\label{tab:question-swap}
\end{table}

\section{Calibration and Provenance-Route Stability}

\begin{figure*}[!tbp]
\centering
\includegraphics[width=\textwidth]{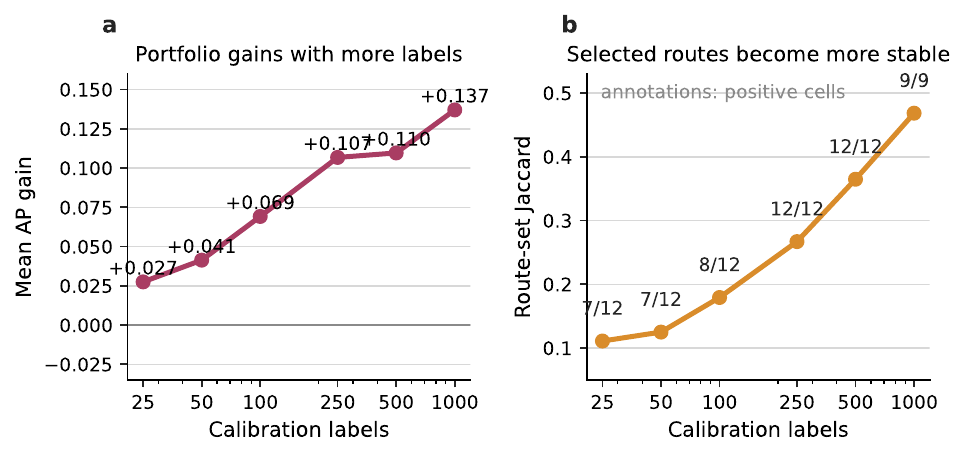}
\caption{Calibration-label efficiency. (a) Mean AP gain over confidence as the number of calibration labels increases. (b) Route-set Jaccard across the same budgets; text labels report the number of positive model--dataset cells. This audit fixes the provenance family.}
\label{fig:supp-label-efficiency}
\end{figure*}

\textbf{Label budget and stability.} Low-label and zero-target-label settings cannot reliably compare families, so this section fixes the provenance family and audits its sparse route selection. Figure~\ref{fig:supp-label-efficiency} and Table~\ref{tab:route-stability} report the amount of calibration data needed and the stability of selected routes across folds. The fallback learns no dense per-route coefficients, but uses validation labels to choose a small route portfolio and one fusion weight.

\begin{table}[H]
\centering
\footnotesize
\setlength{\tabcolsep}{3pt}
\begin{tabular}{lrrrr}
\toprule
Group & Routes & Mean $k$ & Jaccard & Unique \\
\midrule
InternVL3.5-8B & 144 & 17.7 & 0.609 & 27.0 \\
LLaVA-1.5-7B & 128 & 16.4 & 0.668 & 24.2 \\
Qwen3-VL-8B & 144 & 12.7 & 0.424 & 25.8 \\
\midrule
All & 139 & 15.6 & 0.567 & 25.7 \\
\bottomrule
\end{tabular}
\caption{Provenance-route selection stability across held-out folds. Jaccard measures
pairwise overlap between selected route sets; lower values indicate stronger
dataset- or fold-specific calibration.}
\label{tab:route-stability}
\end{table}

\textbf{Final nested-controller seed sensitivity.} Table~\ref{tab:nested-seed-stability} reruns the complete learning algorithm under 50 image-grouped fold seeds: the outer split, inner family comparison, route identities, portfolio size, normalization, and fusion weight are all reselected. This is an algorithm-level stability check, whereas the main bootstrap conditions on the realized out-of-fold fits. The final nested controller retains mean $\Delta$AP $+0.129$ with empirical interval [$+0.126,+0.134$], and 581/600 cell--seed gains are positive. Provenance is selected in 2,094/3,000 outer folds; within those portfolios, question-residual routes occupy 14,462/30,714 selected slots (47.1\%). At the reported seed they occupy 300/617 slots (48.6\%) and 38.9\% of standardized absolute pre-fusion route mass. Thus the global anchor is strong, but the question-specific channel is a substantial selected component rather than an unused add-on.

\begin{table*}[!tbp]
\centering
\footnotesize
\setlength{\tabcolsep}{4pt}
\begin{tabular}{llrrrrr}
\toprule
Model & Dataset & $\Delta$AP mean$\pm$std & Seed 95\% interval & Positive & Prov./conc. folds & Residual slots \\
\midrule
Qwen3-VL-8B & AMBER-D & +0.105$\pm$0.002 & [+0.102,+0.108] & 50/50 & 250/0 & 56.4\% \\
 & Causal-HalBench & +0.095$\pm$0.002 & [+0.089,+0.097] & 50/50 & 12/238 & 56.8\% \\
 & VSR & +0.032$\pm$0.012 & [+0.012,+0.052] & 50/50 & 127/123 & 66.8\% \\
 & HallusionBench & +0.045$\pm$0.009 & [+0.025,+0.061] & 50/50 & 160/90 & 13.3\% \\
\midrule
LLaVA-1.5-7B & AMBER-D & +0.194$\pm$0.004 & [+0.185,+0.199] & 50/50 & 242/8 & 0.6\% \\
 & Causal-HalBench & +0.167$\pm$0.003 & [+0.161,+0.173] & 50/50 & 186/64 & 31.0\% \\
 & VSR & +0.004$\pm$0.011 & [-0.014,+0.025] & 31/50 & 92/158 & 64.0\% \\
 & HallusionBench & +0.076$\pm$0.012 & [+0.050,+0.094] & 50/50 & 215/35 & 48.4\% \\
\midrule
InternVL3.5-8B & AMBER-D & +0.353$\pm$0.003 & [+0.349,+0.355] & 50/50 & 249/1 & 56.9\% \\
 & Causal-HalBench & +0.271$\pm$0.003 & [+0.263,+0.274] & 50/50 & 249/1 & 34.6\% \\
 & VSR & +0.147$\pm$0.008 & [+0.131,+0.161] & 50/50 & 204/46 & 57.0\% \\
 & HallusionBench & +0.062$\pm$0.010 & [+0.044,+0.085] & 50/50 & 108/142 & 34.7\% \\
\midrule
\textbf{Macro} & 12 cells & \textbf{+0.129$\pm$0.002} & \textbf{[+0.126,+0.134]} & \textbf{581/600} & 2094/906 & 47.1\% \\
\bottomrule
\end{tabular}
\caption{Final nested WEP stability under complete grouped-fold reselection.
Each of 50 seeds reruns family choice, route and portfolio-size selection,
normalization, and confidence fusion. Prov./conc. counts the selected family
over 3,000 outer folds. Residual slots are the seed mean for each cell and the
pooled fraction for the macro row.}
\label{tab:nested-seed-stability}
\end{table*}

\textbf{Fusion-direction audit.} The search grid permits either sign so the validation protocol does not encode the desired conclusion. All 60 final outer-fold selections choose $\beta>0$: higher evidence risk is consistently combined with, rather than used to reverse, confidence risk.

\section{Transfer and Sanity Checks}

\begin{figure}[H]
\centering
\includegraphics[width=\linewidth]{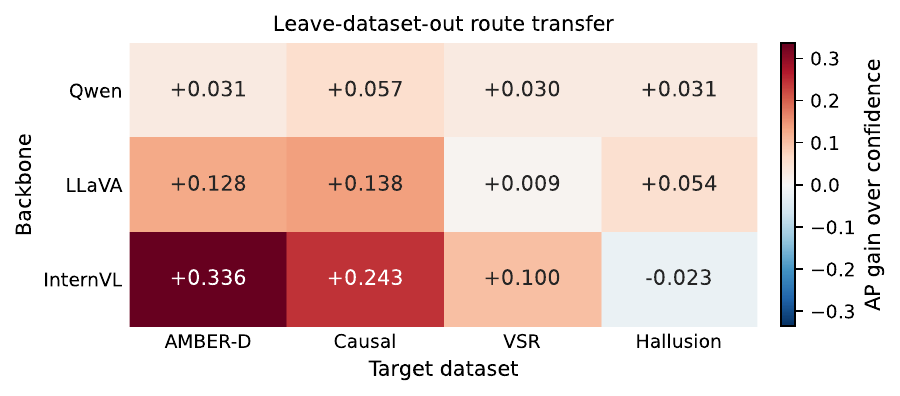}
\caption{Zero-target-label leave-dataset-out transfer. Each cell reports AP gain of the predefined provenance family over confidence when the target dataset is not used for route selection, normalization, or fusion-weight selection.}
\label{fig:supp-transfer-heatmap}
\end{figure}

\textbf{Leave-dataset-out provenance transfer.} Figure~\ref{fig:supp-transfer-heatmap} gives the zero-target-label leave-dataset-out transfer results. The target dataset is excluded from route selection, route normalization, and fusion-weight selection, so the figure tests whether the predefined route family remains useful beyond the dataset on which it was chosen. Table~\ref{tab:transfer-uncertainty} adds a paired image-cluster interval: macro $\Delta$AP is $+0.095$ [$+0.085,+0.102$], with 11/12 positive targets and 9/12 target intervals strictly above zero. InternVL--HallusionBench is the only negative point estimate. We additionally hash the exact evaluation manifests: all 4,331 unique images are readable, and all six dataset pairs have zero shared SHA-256 image hashes. The stricter image--question identity check, resolved paths, and basenames also have zero overlap.

\begin{table}[H]
\centering
\scriptsize
\setlength{\tabcolsep}{2pt}
\begin{tabular}{@{}p{0.38\columnwidth}rr@{}}
\toprule
Transfer comparison & Mean $\Delta$AP [95\% CI] & Positive \\
\midrule
Provenance fallback $-$ confidence & +0.095 [+0.085, +0.102] & 11/12 \\
\bottomrule
\end{tabular}
\caption{Paired uncertainty for same-model leave-dataset-out transfer. Target labels, normalization statistics, route selection, and fusion selection are excluded. The interval macro-averages 1,000 paired image-cluster bootstrap replicates across 12 targets.}
\label{tab:transfer-uncertainty}
\end{table}

\begin{figure*}[!tbp]
\centering
\includegraphics[width=0.95\textwidth]{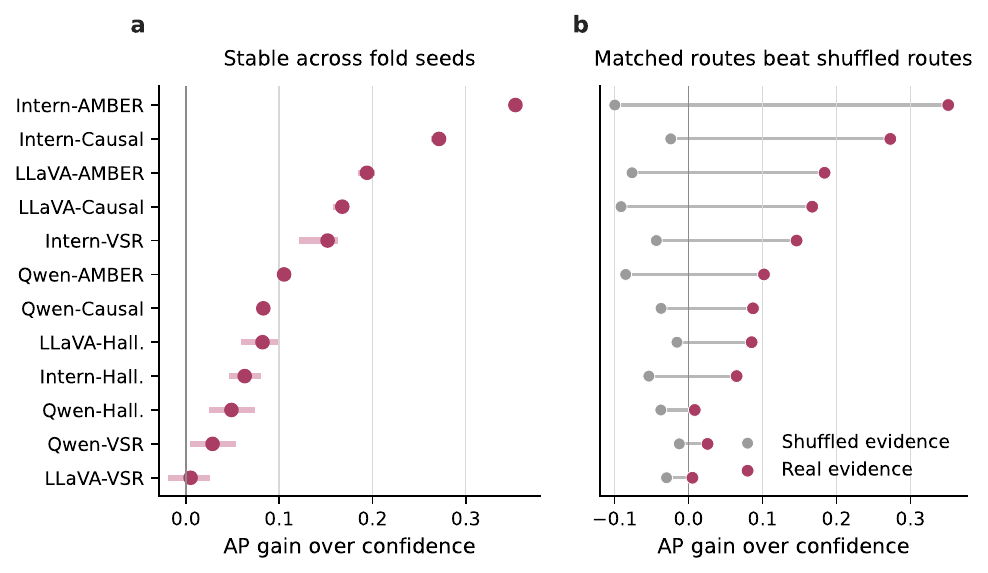}
\caption{Provenance-family robustness and evidence-sanity diagnostics. (a) Seed sensitivity over 50 image-grouped fold seeds; horizontal ranges show the observed seed range. (b) Matched versus shuffled evidence with confidence and labels fixed. Both panels use the row ordering shown in (a).}
\label{fig:supp-robustness-sanity}
\end{figure*}

\textbf{Provenance evidence and label nulls.} Figure~\ref{fig:supp-robustness-sanity}b keeps each example's confidence score and label fixed but shuffles evidence scores across examples, destroying example-specific witness alignment. Table~\ref{tab:label-permutation-null} adds a route-search null that reruns route selection after shuffling labels on each full model--dataset cell. Together, these controls test whether the provenance gain comes from matched evidence rather than a generic score-scale shift or route-search freedom. Across 50 trials per cell, the real mean gain is $+0.130$ AP while the null mean is approximately zero. Eleven of 12 cells have empirical $p\leq0.039$, including nine at the plus-one resolution floor of $1/51$; LLaVA--VSR is the exception, consistent with its small observed gain and weak separation from the permutation null.

\begin{table}[H]
\centering
\small
\setlength{\tabcolsep}{4pt}
\begin{tabular}{lc}
\toprule
Label-permutation route-selection null & Value \\
\midrule
Cells & 12 \\
Null trials per cell & 50 \\
Rows per cell & All \\
Mean sampled rows & 6462.0 \\
Real mean $\Delta$AP & +0.130 \\
Null mean $\Delta$AP & -0.000 \\
Null max $\Delta$AP, mean over cells & +0.036 \\
Cells with positive null mean & 4/12 \\
Cells where null max reaches real gain & 3/12 \\
Cells with empirical $p\leq0.05$ & 11/12 \\
Cells at resolution floor (0.020) & 9/12 \\
Largest empirical $p$ (LLaVA-VSR) & 0.647 \\
\bottomrule
\end{tabular}
\caption{Label-permutation null with route reselection. Each null trial
shuffles benchmark error labels, then reruns the same route, portfolio-size,
and fusion-weight selection used by the provenance fallback with image-grouped folds on each full
cell. The test estimates whether the search space alone can manufacture
gains comparable to the real labels. Empirical $p$ values use plus-one correction.}
\label{tab:label-permutation-null}
\end{table}

\section{Baselines and Interventions}

Unless a table explicitly states nested family selection, matched-budget, transfer, and intervention controls in this section fix the provenance family so every comparator reuses the same cached examples and evidence features. The final nested WEP result is reported in the main paper and Table~\ref{tab:app-full-main}.

\textbf{Baseline families.} Table~\ref{tab:app-perturbation-baselines} adds VCD-margin and ICD-margin, which require an additional contrastive image or prompt path. They are useful comparators, but they answer a different deployment question from WEP's single-prefill risk score.

\begin{table*}[!tbp]
\centering
\small
\setlength{\tabcolsep}{2pt}
\begin{tabular}{@{}p{0.12\textwidth}p{0.11\textwidth}rrrccccc@{}}
\toprule
Model & Dataset & n & Match & Confidence & VCD-margin & ICD-margin & Best VCD & WEP & AP winner \\
\midrule
Qwen3-VL-8B & AMBER-D & 13656 & 0.96 & 0.270/0.705 & 0.140/0.476 & 0.290/0.737 & 0.215/0.655 & 0.415/0.800 & WEP \\
Qwen3-VL-8B & Causal-HalBench & 9180 & 0.95 & 0.335/0.756 & 0.226/0.593 & 0.392/0.722 & 0.279/0.724 & 0.445/0.783 & WEP \\
Qwen3-VL-8B & VSR & 1176 & 0.94 & 0.376/0.729 & 0.288/0.607 & 0.283/0.655 & 0.345/0.701 & 0.394/0.732 & WEP \\
LLaVA-1.5-7B & AMBER-D & 11282 & 0.79 & 0.382/0.676 & 0.288/0.498 & 0.180/0.290 & 0.361/0.635 & 0.687/0.848 & WEP \\
LLaVA-1.5-7B & Causal-HalBench & 5660 & 0.58 & 0.241/0.610 & 0.167/0.491 & 0.148/0.446 & 0.197/0.569 & 0.512/0.804 & WEP \\
LLaVA-1.5-7B & VSR & 1102 & 0.88 & 0.463/0.673 & 0.353/0.538 & 0.342/0.542 & 0.411/0.615 & 0.486/0.687 & WEP \\
InternVL3.5-8B & AMBER-D & 12714 & 0.89 & 0.553/0.767 & 0.387/0.707 & 0.184/0.450 & 0.542/0.772 & 0.686/0.874 & WEP \\
InternVL3.5-8B & Causal-HalBench & 9039 & 0.93 & 0.706/0.798 & 0.786/0.866 & 0.382/0.507 & 0.835/0.899 & 0.874/0.918 & WEP \\
InternVL3.5-8B & VSR & 1096 & 0.88 & 0.639/0.792 & 0.418/0.659 & 0.260/0.403 & 0.573/0.771 & 0.609/0.785 & Confidence \\
\midrule
\multicolumn{2}{l}{Mean $\Delta$AP vs. confidence} & -- & -- & 0.000 & -0.101 & -0.167 & -0.023 & +0.127 & -- \\
\bottomrule
\end{tabular}
\caption{Extra-path contrastive controls versus WEP on the same-prediction subset.
Entries report error AP/AUROC. The Match column reports the fraction of contrastive
records whose predicted answer and correctness label match the WEP feature record.
Best VCD is a diagnostic sweep over VCD-style contrastive weights.}
\label{tab:app-perturbation-baselines}
\end{table*}

\textbf{Readout fidelity, end-to-end interventions, and witness localization.} Before intervention tests, Table~\ref{tab:readout-fidelity} audits the source-token estimator itself. Let $\Delta x_{\ell}^{\rm attn}$ be the actual bias-free attention output at the decision position. We compare $\sum_s C_{\ell,y}(s)$ with $\langle\tilde d_y,\Delta x_{\ell}^{\rm attn}\rangle$ (closure), then compare that linear readout with the exact local change obtained by adding the same update to the layer input and reevaluating the candidate margin through the final RMSNorm and unembedding. The median closure error is below 0.1\% for all backbones, while rank correlation is at least 0.975 and sign agreement is 94.6--97.2\%. Thus the source-token equation faithfully decomposes the local attention update. The separate patching tests below intervene inside the model, execute all remaining blocks, and assess effects on the final output margin.

\begin{table}[!tbp]
\centering
\small
\setlength{\tabcolsep}{3.2pt}
\resizebox{\columnwidth}{!}{%
\begin{tabular}{lrrrr}
\toprule
Backbone & Closure err. $\downarrow$ & Spearman $\rho$ $\uparrow$ &
Sign agree. $\uparrow$ & Local err. $\downarrow$ \\
\midrule
Qwen3-VL-8B & 0.08 & 0.994 & 97.2 & 7.27 \\
LLaVA-1.5-7B & 0.02 & 0.975 & 94.6 & 11.04 \\
InternVL3.5-8B & 0.07 & 0.993 & 95.9 & 7.49 \\
\bottomrule
\end{tabular}
}
\caption{Local fidelity of the source-token readout on the same 64 AMBER-D
examples for each backbone, using every language-model layer. Closure and
local errors are median symmetric absolute errors (\%); $\rho$ and sign
agreement compare the linear readout with the exact candidate-margin change
after adding the same attention update to the layer input. Metrics use the
model-predicted candidate over 2,048--2,304 example--layer observations.}
\label{tab:readout-fidelity}
\end{table}

Table~\ref{tab:cross-backbone-matched-patching} reports end-to-end candidate-margin interventions: selected witness tokens are patched at the same layer used to score evidence, after which the unmodified remaining blocks produce the final output logits. The resulting final-margin change is compared with matched controls. Table~\ref{tab:intervention-traceability} makes the three intervention effects quoted in the main paper directly traceable to their protocol, sample size, and paired interval. Significant effects in 13 of 18 fixed layer--dataset tests, including both positive and reversed signs, show that the readout-selected tokens remain consequential downstream while their net effect is depth dependent. These checks establish intervention relevance, not a path-specific causal decomposition of every intervening block. Table~\ref{tab:witness-localization} separately audits witness maps from all three backbones against Grounded-VSR boxes. This external check is not part of WEP scoring; it tests whether question-bound maps concentrate on the named visual entities relative to area-matched random and uniform references. Pointing accuracy asks whether the maximum-witness token center falls inside the target box; witness mass sums probability inside the box.

\begin{table*}[!tbp]
\centering
\small
\setlength{\tabcolsep}{4pt}
\begin{tabular}{lllrrl}
\toprule
Backbone / dataset & Intervention contrast & Layer / $k$ & $n$ & Gap & 95\% CI \\
\midrule
Qwen / VSR & selected witnesses $-$ random visual tokens & all / 3 & 1249 & +0.397 & [+0.260, +0.538] \\
Qwen / Causal-HalBench & supportive $-$ contradictory tokens & 30 / 3 & 100 & +0.043 & [+0.023, +0.064] \\
Qwen / HallusionBench & selected supportive $-$ random tokens & 35 / 3 & 100 & +0.022 & [+0.008, +0.034] \\
\bottomrule
\end{tabular}
\caption{Traceability of the three intervention effects summarized in the main
paper. The intervention is inserted at the indicated layer, all remaining
blocks are executed, and gaps are paired changes in the final predicted-candidate
margin. VSR uses hidden ablation over the full evaluation set; Causal-HalBench
and HallusionBench use value patching. Intervals are paired image-cluster
bootstrap intervals.}
\label{tab:intervention-traceability}
\end{table*}

\begin{table*}[!tbp]
\centering\small
\begin{tabular}{lcccc}
\toprule
Backbone & Subject point/random & Object point/random & Subject mass/uniform & Object mass/uniform \\
\midrule
Qwen3-VL-8B & 0.699/0.361 & 0.707/0.393 & 0.602/0.353 & 0.607/0.384 \\
LLaVA-1.5-7B & 0.587/0.355 & 0.538/0.387 & 0.574/0.353 & 0.581/0.384 \\
InternVL3.5-8B & 0.776/0.355 & 0.706/0.386 & 0.650/0.353 & 0.617/0.384 \\
\bottomrule
\end{tabular}
\caption{Question-bound witness localization on VSR boxes. ``Point'' is the
fraction of examples whose maximum-witness token center lies inside the target
box, compared with the box-area random expectation. ``Mass'' is witness
probability inside the box, compared with a uniform visual-token distribution.}
\label{tab:witness-localization}
\end{table*}

\begin{table*}[!tbp]
\centering\small
\setlength{\tabcolsep}{5pt}
\begin{tabular}{llccc}
\toprule
Backbone & Dataset & Early layer & Middle layer & Late layer \\
\midrule
Qwen3-VL-8B & AMBER-D & L24: -0.022$^{-}$ & L30: +0.039$^{+}$ & L35: +0.037$^{+}$ \\
Qwen3-VL-8B & Causal-HalBench & L24: -0.026$^{-}$ & L30: +0.043$^{+}$ & L35: +0.016 \\
LLaVA-1.5-7B & AMBER-D & L21: +0.001$^{+}$ & L26: +0.003$^{+}$ & L31: +0.002$^{+}$ \\
LLaVA-1.5-7B & Causal-HalBench & L21: +0.000 & L26: +0.005$^{+}$ & L31: +0.005$^{+}$ \\
InternVL3.5-8B & AMBER-D & L24: -0.048$^{-}$ & L30: -0.030$^{-}$ & L35: +0.010$^{+}$ \\
InternVL3.5-8B & Causal-HalBench & L24: +0.004 & L30: +0.012 & L35: +0.004 \\
\bottomrule
\end{tabular}
\caption{End-to-end value patching on 100 examples per cell at three depths
fixed before inspection. We patch the top three visual tokens at the layer used
for evidence scoring, execute all remaining blocks, and report the
supportive-minus-contradictory gap in the final output margin. Superscripts mark
95\% image-cluster bootstrap intervals strictly above ($^+$) or below ($^-$)
zero; 13/18 fixed tests are nonzero. Reversed signs at some depths motivate
layer-specific selection rather than uniform aggregation.}
\label{tab:cross-backbone-matched-patching}
\end{table*}

\begin{figure*}[!tbp]
\centering
\includegraphics[width=\textwidth]{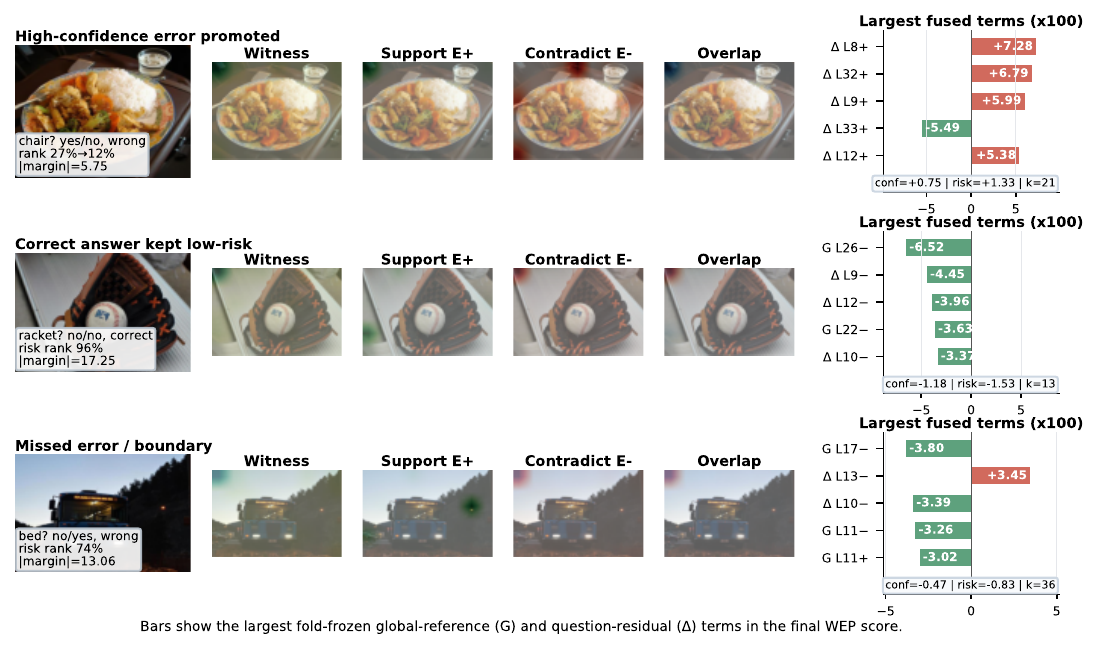}
\caption{Qualitative evidence and route contributions. Stored provenance cases show the input, witness, signed evidence, overlap, and dominant routes for a promoted error, a retained correct answer, and a representative miss; green/red bars lower/raise risk. Concentration reuses the signed evidence through normalized entropy.}
\label{fig:app-evidence-maps}
\end{figure*}

\textbf{Direct sparse provenance-route learners.} Table~\ref{tab:sparse-route-baselines} addresses whether conventional sparse supervision can replace the unweighted top-$k$ provenance portfolio. Both controls keep confidence as a fixed-direction anchor and learn only a route residual; L1 strength and fusion are selected on inner image-group folds. With full labels, L1 is a stronger in-domain upper bound but uses 80.2 weighted routes on average. Stability selection reduces this to 28.5 routes and nearly matches the provenance portfolio, which uses 15.0 selected routes and one fusion weight, and transfers better without target labels ($+0.095$ versus $+0.073/+0.070$ mean AP gain). Thus unrestricted sparse weights improve target-domain fit, whereas the portfolio offers a smaller auditable monitor with stronger transfer. At 100 and 250 labels, L1 and the provenance portfolio have similar mean gains; the portfolio is positive on two more cells at 250 labels.

\begin{table}[H]
\centering
\scriptsize
\setlength{\tabcolsep}{1.2pt}
\begin{tabular}{lrrrr}
\toprule
Detector & In-domain AP & $\Delta$AP & Routes & Transfer $\Delta$AP \\
\midrule
Confidence & 0.468 & +0.000 & 0 & +0.000 \\
L1 route LR & 0.641 & +0.172 & 80.2 & +0.073 \\
Stability-L1 & 0.601 & +0.133 & 28.5 & +0.070 \\
Provenance portfolio & 0.593 & +0.125 & 15.0 & +0.095 \\
\bottomrule
\end{tabular}
\caption{Direct sparse supervised controls over the same confidence and
layer-signed route features. L1 learns a route residual while preserving
confidence as a fixed-direction anchor; strength and fusion are chosen on inner image-group folds.
Stability-L1 retains routes repeatedly selected across calibration subsamples.
Transfer trains on the other three datasets of the same backbone without target
labels. Route counts exclude confidence and are averaged over cells/folds.}
\label{tab:sparse-route-baselines}
\end{table}

\noindent\textbf{Full-vocabulary confidence controls.} Table~\ref{tab:vocab-baselines} tests whether candidate margin is weak only because it discards noncandidate vocabulary logits. From the same original prefill, we retain candidate NLL, entropy, energy, top-1 uncertainty, and the predicted-candidate margin to the strongest noncandidate token. We also fit a six-feature LR using these five scalars and candidate confidence under the same image-grouped out-of-fold protocol. Predictions agree with the cached provenance runs on 99.1--100\% of each cell; the few numerical boundary flips are excluded from all detectors. Candidate margin reaches $0.463$ AP, the best vocabulary scalar reaches $0.434$, and their LR fusion reaches $0.403$, versus $0.540$ for the provenance portfolio on identical rows. Thus omitted full-vocabulary uncertainty does not explain the witness-route gain.

\begin{table}[H]
\centering\footnotesize
\setlength{\tabcolsep}{3pt}
\begin{tabular}{lrrr}
\toprule
Single-prefill detector & AP $\uparrow$ & $\Delta$AP & Pos. \\
\midrule
Candidate margin & 0.463 & +0.000 & 0/12 \\
Full-vocab candidate NLL & 0.324 & -0.139 & 1/12 \\
Full-vocab entropy & 0.434 & -0.029 & 3/12 \\
Full-vocab energy & 0.417 & -0.045 & 2/12 \\
Cand.--best noncand. margin & 0.312 & -0.151 & 1/12 \\
Full-vocab top-1 uncertainty & 0.413 & -0.049 & 2/12 \\
\midrule
Vocab-summary+conf. LR & 0.403 & -0.060 & 4/12 \\
\midrule
\textbf{Provenance fallback} & 0.540 & +0.078 & 10/12 \\
\bottomrule
\end{tabular}
\caption{Full-vocabulary confidence controls under the same image-grouped evaluation. LR uses candidate confidence plus five compact vocabulary scalars; no full logit vectors are retained. Each cell uses a deterministic subset of at most 2,000 examples, with every detector compared on identical same-prediction rows.}
\label{tab:vocab-baselines}
\end{table}

Candidate-adapted Overthinking and trajectory controls reach 0.334 and 0.528 AP, respectively, under the same grouped subsets and labels. Stronger learned selectors over all-layer decision and pooled multimodal states are analyzed below alongside the closest published detector.

Table~\ref{tab:halp-baseline} gives the closest published hallucination detector under our closed-answer protocol. HALP extracts pre-generation visual or query-token states and trains a 512--256--128 MLP (Kogilathota et al., 2026). We use its official query-token architecture, dropout, optimizer, batch size, and 50-epoch schedule. The $3L/4$ branch is fixed before fitting: HALP reports that query tokens dominate on most tested models and identifies this depth as optimal for its Qwen2.5-VL and LLaVA-Next backbones. We replace its random 80/20 split with our stricter image-grouped five-fold split and repeat three fitting seeds. HALP-QT is a strong in-domain detector, reaching $0.610\pm0.003$ AP, but the provenance-minus-HALP gap of $-0.018$ has a paired 95\% interval of $[-0.036,+0.001]$. It transfers below confidence by $0.065$ AP on average, while provenance remains $0.089$ above confidence on the same subsets. Appending confidence to HALP does not change this conclusion.

\begin{table*}[!tbp]
\centering
\scriptsize
\setlength{\tabcolsep}{3.2pt}
\begin{tabular}{lrrrrrrr}
\toprule
Detector & Params & AP$\uparrow$ & AUROC$\uparrow$ & AURC$\downarrow$ & Prov. wins & Transfer $\Delta$AP & Pos. \\
\midrule
Confidence & 0 & 0.473 & 0.687 & 0.217 & -- & $0.000$ & -- \\
Provenance fallback & 1 wt. & 0.592 & 0.762 & 0.184 & -- & $+0.089$ & 10/12 \\
HALP-QT@$3L/4$ & 2.26M & $0.610\pm0.003$ & $0.759\pm0.001$ & $0.187\pm0.001$ & 7/12 & $-0.065\pm0.006$ & 2.3/12 \\
HALP-QT@$3L/4$+conf. & 2.26M & $0.609\pm0.011$ & $0.762\pm0.005$ & $0.184\pm0.002$ & 7/12 & $-0.064\pm0.011$ & 3.0/12 \\
\bottomrule
\end{tabular}
\caption{Matched-protocol adaptation of HALP (Kogilathota et al., 2026). We use its published query-token MLP architecture and training hyperparameters at a fixed normalized $3L/4$ depth, replacing HALP's random 80/20 split with the same image-grouped folds and labels as the provenance fallback. The confidence variant appends one standardized candidate-margin feature. HALP values are mean$\pm$standard deviation over three fitting seeds. ``Prov. wins'' counts cells where provenance has higher AP; ``Pos.'' is the mean positive-target count over seeds. Transfer uses no target labels. Provenance values are recomputed on these same matched subsets.}
\label{tab:halp-baseline}
\end{table*}

Table~\ref{tab:hidden-mlp-robustness} tests whether that comparison depends on one projection width or MLP initialization. It varies 4, 8, and 16 projected coordinates per decoder layer and repeats each MLP over five fitting seeds while keeping the image-group folds fixed. The 16-coordinate MLP reaches $0.594\pm0.004$ AP, close to provenance's $0.592$ within fitting-seed variation. Paired common-row resampling in Table~\ref{tab:hidden-mlp-robustness} gives a provenance-minus-MLP difference of $-0.002$ [$-0.020,+0.015$]. The methods are therefore statistically indistinguishable in macro in-domain AP. In the same-model leave-dataset-out test on these deterministic matched-budget subsets, each detector is fit on the other three datasets and receives no labels, normalization statistics, or model-selection signal from the target. All three MLP widths fall below confidence on average; the strongest gives $-0.058\pm0.014$ mean $\Delta$AP, whereas provenance gives $+0.089$ and is positive on 10/12 targets. Thus the main advantage over the dense probe is sparse transfer behavior and auditability, not a large in-domain AP gap.

\begin{table*}[!tbp]
\centering
\small
\setlength{\tabcolsep}{4.5pt}
\begin{tabular}{lrrrrr}
\toprule
Detector & Params & In-domain AP & Prov. wins & Transfer $\Delta$AP & Pos. \\
\midrule
MLP, 4/layer & 9.4--10.4k & $0.562\pm0.002$ & 8/12 & $-0.086\pm0.011$ & 0/12 \\
MLP, 8/layer & 17.6--19.6k & $0.587\pm0.004$ & 7/12 & $-0.071\pm0.013$ & 2/12 \\
MLP, 16/layer & 34.0--38.0k & $0.594\pm0.004$ & 7/12 & $-0.058\pm0.014$ & 1/12 \\
Multimodal, 6/stream & 38.0--42.7k & $0.551\pm0.006$ & 8/12 & $-0.074\pm0.007$ & 1/12 \\
Full-state LR & 12.3k & 0.542 & 8/12 & $-0.070$ & 3/12 \\
Provenance fallback & sparse & 0.592 & -- & +0.089 & 10/12 \\
\bottomrule
\end{tabular}
\caption{Learned-selector robustness over five fitting seeds with fixed image-group
folds. Decision-state MLP entries vary the number of cached label-independent
projections per layer; the multimodal MLP uses six pooled image, question, and
decision coordinates per stream and layer. MLP values are macro mean $\pm$
standard deviation across seeds. The full-state LR uses unprojected pooled
decision, image, and question states at an inner-selected layer with one fixed
fitting seed.
Transfer trains on the other three datasets from the same backbone without
target labels and is evaluated on the same deterministic per-cell subsets.
``Prov. wins'' compares cell-wise provenance AP with each detector. For the
16-coordinate MLP, the paired provenance-minus-MLP AP difference is
$-0.002$ [$-0.020,+0.015$] under 1,000 image-cluster bootstrap replicates.}
\label{tab:hidden-mlp-robustness}
\end{table*}

The final MLP row in Table~\ref{tab:hidden-mlp-robustness} strengthens this control by adding explicit pooled image and question states to every layer's projected decision state and confidence. The primary width of six coordinates per stream and layer was fixed before evaluation; widths four and eight are sensitivity checks. The primary 64--16 MLP has 38.0k--42.7k parameters and uses exactly the same labels, frozen image-group folds, and transfer protocol as the decision-state selector. It improves in-domain AP over confidence from $0.473$ to $0.551\pm0.006$, but is worse than provenance and transfers below confidence by $0.074\pm0.007$ AP, with only 1/12 positive targets. Thus merely exposing a generic classifier to pooled multimodal states does not reproduce the sparse witness-route signal.

The full-state LR row in Table~\ref{tab:hidden-mlp-robustness} removes the fixed random projection entirely. At four prespecified normalized depths, it concatenates confidence with the unprojected 4096-dimensional pooled decision, image, and question states and fits a 12.3k-parameter linear logistic probe. The layer and L2 strength are selected on an inner image-group split; outer folds and deterministic subsets match the provenance route signal. This stronger input raises AP from $0.473$ to $0.542$, but provenance reaches $0.592$, wins 8/12 cells, and also has higher AUROC and lower AURC. Without target labels, the full-state probe gives $-0.070$ mean $\Delta$AP with 3/12 positive targets, versus provenance's $+0.089$ and 10/12.

\section{Deployment and Boundary Analysis}

\textbf{Latency.} Table~\ref{tab:latency} reports the measured overhead of the current Python-hook implementation. The table supports the path-count claim used in the main paper--no extra model forward, crop, perturbation, backward pass, or external verifier--but it should not be read as an optimized deployment latency bound. Measurements use one NVIDIA A800 80GB GPU, PyTorch 2.6.0 with CUDA 12.4, Transformers 4.57.1, batch size one, explicit CUDA synchronization, and four paired warm-up examples before 60 measured probes. Qwen uses SDPA with automatic dtype, LLaVA uses eager attention in FP16, and InternVL uses BF16. The absolute increment is $0.050$--$0.107$ seconds per example. Both route families reuse this extraction; family selection and compact scalar scoring add negligible postprocessing relative to the hooks. Relative overhead falls to 28.9\% for the high-token InternVL mp12 setting. Compared with candidate-only scoring, WEP adds 122/68 MB of peak allocated memory for Qwen/LLaVA and less than 1 MB for either InternVL setting; visual activations dominate the latter peaks.

\begin{table}[H]
\centering
\footnotesize
\setlength{\tabcolsep}{3pt}
\resizebox{\linewidth}{!}{%
\begin{tabular}{lrrrrrr}
\toprule
Model/backend & Base s/ex & WEP s/ex & Over. & WEP ex/s & Base +MB & WEP +MB \\
\midrule
Qwen3-VL-8B & 0.046 & 0.099 & 116.4\% & 10.13 & 14 & 136 \\
LLaVA-1.5-7B & 0.072 & 0.123 & 71.8\% & 8.11 & 124 & 192 \\
InternVL mp4 & 0.056 & 0.106 & 88.3\% & 9.47 & 244 & 244 \\
InternVL mp12 & 0.369 & 0.476 & 28.9\% & 2.10 & 1038 & 1038 \\
\bottomrule
\end{tabular}}
\caption{Measured batch-one latency and incremental peak allocated GPU memory on 60 AMBER probes. Times include the current Python hooks and compact route extraction; they are not an optimized fused-kernel bound.}
\label{tab:latency}
\end{table}

\textbf{Review-queue audit.} Table~\ref{tab:failure-audit} characterizes both sides of selective use: benchmark-correct answers entering the review queue and errors remaining below a given review budget. Some apparent false positives are lucky-correct or ambiguous cases, while the remaining misses motivate using WEP to prioritize a finite review budget rather than as a hard accept/reject rule.

\begin{table}[H]
\centering
\small
\setlength{\tabcolsep}{4pt}
\begin{tabular}{lcc}
\toprule
Audit view & Confidence & WEP \\
\midrule
Top-5\% precision & 53.6\% & 74.2\% \\
Top-5\% false positives & 46.4\% & 25.8\% \\
Top-5\% error recall & 9.6\% & 13.5\% \\
Top-10\% precision & 51.1\% & 69.0\% \\
Top-10\% false positives & 48.9\% & 31.0\% \\
Top-10\% error recall & 18.5\% & 24.6\% \\
Errors beyond 50\% risk & 29.3\% & 22.5\% \\
Unique top-5\% caught errors & 5.1\% & 9.0\% \\
\bottomrule
\end{tabular}
\caption{Failure-audit summary for final nested WEP over the 12 main cells. Precision counts benchmark errors among reviewed examples; false positives are benchmark-correct answers entering the review queue.}
\label{tab:failure-audit}
\end{table}

\textbf{Spatial relations.} VSR is the clearest stress test. Many VSR mistakes are relation errors in which the model has visually accessed both relevant entities but fails to resolve the spatial predicate. WEP measures whether selected evidence routes cover entity witnesses with positive or negative sign; it does not represent geometry, relative position, or directional predicates. This explains why VSR gains are smaller than AMBER-D and Causal-HalBench gains.

Table~\ref{tab:vsr-geometry-diagnostic} then asks whether explicit geometry would help if it were available. Using Grounded-VSR boxes only as an oracle-style diagnostic, we compute a simple subject--object geometry score and combine it with the held-out provenance risk. This raises macro AP from $0.540$ to $0.562$ on all VSR examples and from $0.555$ to $0.581$ on 2D spatial relations. The result supports the failure diagnosis: VSR needs geometric reasoning signals beyond the current witness-route portfolio.

\begin{table}[H]
\centering
\scriptsize
\setlength{\tabcolsep}{3pt}
\begin{tabular}{lrrrrrr}
\toprule
Model & $n$ & Conf. & Prov. & BBox geom. & Prov.+geom. & $\Delta$ \\
\midrule
Qwen & 1249 & 0.373 & 0.381 & 0.286 & 0.397 & +0.016 \\
LLaVA & 1249 & 0.472 & 0.489 & 0.463 & 0.525 & +0.035 \\
InternVL & 1249 & 0.600 & 0.749 & 0.560 & 0.765 & +0.017 \\
Macro & -- & 0.482 & 0.540 & 0.436 & 0.562 & +0.023 \\
\midrule
2D rel. macro & -- & 0.476 & 0.555 & 0.453 & 0.581 & +0.025 \\
\bottomrule
\end{tabular}
\caption{VSR bbox-geometry diagnostic. Grounded-VSR boxes are used only
as an oracle-style diagnostic. BBox geom. scores whether explicit
subject--object geometry contradicts the model prediction; Prov.+geom.
combines held-out provenance risk with this geometry score. The additional
gains, especially on 2D relations, indicate that VSR errors require explicit
spatial geometry beyond routing relation types to existing provenance routes.}
\label{tab:vsr-geometry-diagnostic}
\end{table}

\textbf{Small mixed benchmarks.} HallusionBench combines visual illusion, chart/table reading, and language prior conflicts. The sample size is much smaller than AMBER-D and Causal-HalBench, so fold-level route selection is noisier. WEP remains positive in AP on all three backbones, but confidence intervals are wider.

\textbf{Four-choice candidate pilot.} Table~\ref{tab:mmstar-pilot} evaluates Qwen3-VL-8B on a deterministic, category-stratified 300-example MMStar pilot with candidates \texttt{A/B/C/D}. Each first-level MMStar category contributes 50 examples. We append only an answer-format instruction requesting one option letter and otherwise use the same prefill instrumentation and image-grouped five-fold controller protocol. Candidate confidence obtains 0.575 error AP and WEP 0.586, a modest $+0.011$ gain with interval $[-0.057,+0.070]$. The category slices sharpen the method boundary: gains are largest on fine-grained and coarse perception, while logical and science/technology reasoning are negative. This is consistent with WEP diagnosing visual evidence provenance rather than replacing missing knowledge or multi-step reasoning.

The candidate-set argmax agrees with ordinary greedy generation on 96.7\% of the 300 examples; 99.3\% of generated responses are parseable option letters. Candidate argmax and generation accuracies are 56.0\% and 54.7\%, respectively. Thus the four-choice risk task closely tracks the backbone's standard answer behavior rather than an artificial candidate-only decision.

\begin{table}[!tbp]
\centering
\footnotesize
\setlength{\tabcolsep}{3.2pt}
\begin{tabular}{lrrrr}
\toprule
MMStar slice & $n$ & Conf. AP & WEP AP & $\Delta$AP \\
\midrule
All & 300 & 0.575 & 0.586 & +0.011 \\
Coarse perception & 50 & 0.412 & 0.485 & +0.073 \\
Fine-grained perception & 50 & 0.633 & 0.764 & +0.132 \\
Instance reasoning & 50 & 0.650 & 0.684 & +0.034 \\
Logical reasoning & 50 & 0.570 & 0.477 & -0.092 \\
Math & 50 & 0.710 & 0.771 & +0.061 \\
Science/technology & 50 & 0.567 & 0.487 & -0.080 \\
\bottomrule
\end{tabular}
\caption{Four-choice MMStar pilot with 50 deterministically sampled examples
from each first-level ability category. The overall paired image-bootstrap
interval for $\Delta$AP is $[-0.057,+0.070]$; category rows localize the
observed task heterogeneity.}
\label{tab:mmstar-pilot}
\end{table}

\textbf{Multi-token candidate prototype.} Table~\ref{tab:wep-span} tests whether the evidence construction is restricted to one-token labels. We replace \texttt{yes}/\texttt{no} with two token-aligned, six-token candidate answers and score both from one shared visual prefill plus one batched cache continuation. On all 1,249 VSR examples, WEP-Span improves error AP from 0.383 to 0.541 and error AUROC from 0.691 to 0.770; the paired 95\% intervals for the gains are $[0.110,0.205]$ and $[0.057,0.101]$. Evidence- and label-shuffle nulls do not reproduce the gain. This validates finite multi-token candidate scoring; extending the same construction to unconstrained generation requires claim segmentation and explicit alternatives to rank.

\begin{table}[H]
\centering
\small
\begin{tabular}{lccc}
\toprule
Method & Error AP $\uparrow$ & AUROC $\uparrow$ & P@10\% $\uparrow$ \\
\midrule
Span confidence & 0.383 & 0.691 & 0.464 \\
Evidence only & 0.379 & 0.696 & 0.440 \\
\textbf{WEP-Span} & 0.541 & 0.770 & 0.648 \\
\bottomrule
\end{tabular}
\caption{Provenance-family multi-token VSR prototype with one shared visual
prefill. Each candidate contains six tokens; suffix scores are computed in
one batched text-only cache continuation.}
\label{tab:wep-span}
\end{table}

\end{document}